\documentclass[preprint,3p]{elsarticle}

\usepackage{amssymb}

\usepackage{latexsym,bm}
\usepackage[tbtags]{amsmath}
\usepackage{color}
\usepackage{stfloats}
\usepackage{times}
\usepackage{caption}
\usepackage{amsmath}
\usepackage{amssymb}
\usepackage{amscd}
\usepackage{amsfonts}
\usepackage{graphicx}
\usepackage{psfrag}
\usepackage{epsfig}
\usepackage{algorithm}
\usepackage{algorithmic}
\usepackage{bbm}

\journal{Journal of Information and Intelligence}

\begin{document}

\begin{frontmatter}

\title{Structural Knowledge-Driven Meta-Learning for Task Offloading in Vehicular Networks with Integrated Communications, Sensing and Computing}

%

\author[label1]{Ruijin Sun}
\ead{sunruijin@xidian.edu.cn}
\author[label1]{Yao Wen}
\ead{wenyao@stu.xidian.edu.cn}
\author[label1]{Nan Cheng\corref{cor1}}
\ead{nancheng@xidian.edu.cn}
\author[label1]{Wei Wang}
\ead{weiwang_2042@stu.xidian.edu.cn}
\author[label2]{Rong Chai}
\ead{chairong@cqupt.edu.cn}
\author[label1]{Yilong Hui}
\ead{ylhui@xidian.edu.cn}
\affiliation[label1]{organization={State Key Laboratory of ISN, Xidian University},
            city={Xi'an},
            postcode={710071},
            state={Shanxi},
            country={China}}
\affiliation[label2]{organization={Chongqing Key Laboratory of Mobile Communication Technology, Chongqing University of Posts and Telecommunications},
            city={Chongqing},
            postcode={400065},
            country={China}}
\cortext[cor1]{Corresponding author}

\begin{abstract}
Task offloading is a potential solution to satisfy the strict requirements of computation-intensive and latency-sensitive vehicular applications due to the limited onboard computing resources. However, the overwhelming upload traffic may lead to unacceptable uploading time. To tackle this issue, for tasks taking environmental data as input, the data perceived by roadside units (RSU) equipped with several sensors can be directly exploited for computation, resulting in a novel task offloading paradigm with integrated communications, sensing and computing (I-CSC). With this paradigm, vehicles can select to upload their sensed data to RSUs or transmit computing instructions to RSUs during the offloading. By optimizing the computation mode and network resources, in this paper, we investigate an I-CSC-based task offloading problem to reduce the cost caused by resource consumption while guaranteeing the latency of each task. Although this non-convex problem can be handled by the alternating minimization (AM) algorithm that alternatively minimizes the divided four sub-problems, it leads to high computational complexity and local optimal solution. To tackle this challenge, we propose a creative structural knowledge-driven meta-learning (SKDML) method,  involving both the model-based AM algorithm and neural networks. Specifically, borrowing the iterative structure of the AM algorithm, also referred to as structural knowledge, the proposed SKDML adopts long short-term memory (LSTM) network-based meta-learning to learn an adaptive optimizer for updating variables in each sub-problem, instead of the handcrafted counterpart in the AM algorithm. Furthermore, to pull out the solution from the local optimum, our proposed SKDML updates parameters in LSTM with the global loss function. Simulation results demonstrate that our method outperforms both the AM algorithm and the meta-learning without structural knowledge in terms of both the online processing time and the network performance.


\end{abstract}



\end{frontmatter}

%

\section{Introduction}
\subsection{Motivation and Contribution}
Recently, with the development of intelligent transportation and wireless communications,  vehicular networks have attracted increasing interest from both academia and industry \cite{8767077}. By interconnecting vehicles and infrastructures, such as roadside units (RSUs), vehicular networks extend the range of information exchange, leading to improved transportation efficiency and safety \cite{8744265}, \cite{8936542}, \cite{messous2017computation}. Furthermore, to assist automotive driving, more and more sensors (e.g., cameras, radar) are integrated into vehicles to sense environmental information from all directions, which will generate approximately 1-GB data per second and should be processed by the on-board processors in real-time \cite{8984345}. Due to the limited computing resources on vehicles, locally processing such computation-sensitive tasks cannot meet the latency requirements. One potential solution to significantly lessen the on-broad computational workload and processing latency is the mobile edge computing (MEC) technology, which offloads the sensed environmental data of vehicles to nearby RSUs with edge servers for computing \cite{ren2019survey}, \cite{li2020energy}, \cite{wang2022joint}. 

To jointly utilize the computation resource in edge serves and the communication resource in wireless networks within the integrated communication and computing (I-CC) framework, resource management for task offloading in MEC-enabled vehicular networks has been a hot research topic. Most works in this field aim to reduce the overall processing latency of tasks \cite{zhang2022federated} or the system cost caused by resource consumption \cite{8471165}, as the response time is the primary metric for real-time vehicular applications and resources in networks are scarce and limited. 
In those works \cite{8555636}, a universal phenomenon is gradually revealed that the uploading time of input data is the major source of the latency, due to the limited bandwidth and the large size of input data. With the explosive proliferation of various in-vehicle applications, this dilemma of unaffordable uploading time would become more severe. For example, tasks involving three-dimensional (3-D) reconstruction necessitate the transmission of original high-resolution video frames to RSUs for deep map fusion. Given the substantial volume of video data, like that from a camera boasting 7680$\times$4320 resolution (i.e., 8K resolution) demanding 11.9 Gb/s per pixel at 30 frames per second, attempting to upload within the 10 Gb/s peak rate of a 5G network would take approximately 1.2 seconds\cite{Dame_2013_CVPR}. Such a huge volume of input data results in unacceptable transmission latency for latency-sensitive vehicular networks.

To tackle this challenge, for most driving-related vehicular applications taking environmental data as input, a novel task offloading paradigm with integrated communication, sensing and computing (I-CSC) has emerged \cite{9439524} to exploit not only the computation resource of RSUs with MEC serves but also the environmental data perceived by sensors in RSUs. Specifically, to assist road monitoring and automotive driving, various sensors, such as light detection and ranging (LiDAR), cameras, have been equipped on RSUs, which can acquire similar environmental information with nearby vehicles. Although the data sensed by vehicles and RSUs in different locations provide distinct viewpoints, this can be eliminated by pre-conducting the coordinate transformation with several matrix operations \cite{5209891}. Consequently, the environmental data sensing of RSUs makes computation instruction transmission become a new MEC mode for task offloading. Compared with the data uploading MEC mode, the size of computation instruction will be much smaller, leading to considerably reduced transmission latency. With this I-CSC-based task offloading paradigm, computation mode selection problems are investigated in \cite{9439524} and \cite{9982429} to minimize the offloading latency, where model-based optimization theory and data-driven reinforcement learning are employed, respectively. However, the model-driven approaches leverage mathematical theories like optimization and game theory, and
require precise modeling of dynamic network features, which perform poorly in complex scenarios. Moreover, these approaches involve multiple iterations for real-time computation, leading to longer processing time and unsuitability for low-latency services. On the other hand, the data-driven approaches utilize
neural networks to learn complex mapping relationships from data. They trade offline training for quicker online computation but rely on stable networks and abundant high-quality training data, resulting in poor generalization. Furthermore, data-driven neural networks are regarded as ``black-box" and lack interpretability. 

Motivated by this, in this paper, we investigate an I-CSC-based task offloading problem in vehicular networks, and propose a novel structural knowledge-driven meta-learning (SKDML) method, exploiting both the interpretability of model-based methods and the fast inference of data-driven methods.  Specifically, to figure out the tradeoff between latency and resource consumption in the I-CSC paradigm, in this paper, a joint computation mode selection and resource allocation problem is formulated to minimize the system total cost caused by resource consumption while ensuring the latency requirement of each task, where three computation modes are considered, i.e., the local computation mode, the data transmission-based MEC mode and the instruction transmission-based MEC mode. Then, to solve this non-convex problem, a creative SKDML method is proposed, which keeps the inner-and-outer-iterative structure of the model-based alternative minimization (AM) algorithm, referred to as structural knowledge and adopts long short-term memory (LSTM) networks to learn adaptive strategies for variable updating.  The main contributions of this paper are summarized as follows.

\begin{itemize}

\item This paper investigates resource allocation with a pioneering I-CSC-based computational offloading scheme. While prevailing research predominantly emphasizes latency as the primary optimization criterion, offloading costs are often overlooked. It is crucial to note that as latency diminishes, the associated overhead costs tend to surge. Recognizing cost as a paramount metric, this paper aims to strike a balance between task offloading latency and its associated costs. Consequently, we formulate a problem model that incorporates latency tolerance as a constraint, with the offloading cost as the primary optimization objective.
\item In order to address the challenges posed by the non-convex problem's high computational complexity and susceptibility to locally optimal solutions, the paper presents a groundbreaking structural SKDML method. This method synergistically combines the model-based AM algorithm with neural networks. The SKDML leverages the iterative structure of the AM algorithm, employing LSTM network-based meta-learning to develop an adaptive optimizer tailored for variable updates across individual sub-problems. In addition, the proposed SKDML method is engineered to navigate away from local optima, enhancing solution robustness.


\item Simulation results have shown that the proposed SKDML has a relatively shorter online processing time and superior performance compared to the AM algorithm and the meta-learning approach without knowledge. Specifically, the proposed SKDML has a convergence time improvement of approximately 15\% compared to the AM algorithm, and an improvement of approximately 47\% compared to meta-learning without knowledge. In terms of performance, our proposed algorithm improves by approximately 50\% compared to the AM algorithm and by approximately 47\% compared to the meta-learning without knowledge.
\end{itemize}

\subsection{Related Works}
In this subsection, We introduced the existing model-driven solutions to task offloading problems in the I-CC mode. However, due to the drawbacks of model-driven approaches such as long online processing time, we also introduced the existing paper on data-driven solutions to task offloading problems. Finally, we introduced the tasks offloading issues that exist in I-CSC mode, and summarized our methods.

First, we have introduced existing research papers on mathematical model-driven solutions for tasks offloading problems in I-CC mode. For mathematical model-driven resource allocation methods, Zhang \textit{et al.} \cite{zhang2017optimal} proposed a Stackelberg game-based approach to maximize the utility of both vehicles and MEC servers. Similarly, Zhao \textit{et al.} \cite{zhao2022adaptive} introduced an adaptive vehicle clustering algorithm based on the fuzzy C-means algorithm, which can reduce vehicle power consumption while meeting the required vehicle latency. Liu \textit{et al.} \cite{8422240} propose a distributed computation offloading scheme by formulating the computation offloading decision-making problem as a multi-user game.  Shahzad \textit{et al.} \cite{7726790} used the “dynamic programming with Hamming distance termination” method to offload tasks and reduce the energy use of the mobile device. However, it is more used for noncritical tasks, not suitable for sensor-driven autonomous driving services. Du \textit{et al.} \cite{8543658} devised a continuous relaxation and Lagrangian dual decomposition-based iterative algorithm for joint radio resource and power allocation.  Liu \textit{et al.} \cite{8166725} proposed a game-based distributed computation scheme, where the users compete for the edge cloud’s finite computation resources via a pricing approach. To minimize both latency and energy consumption, Dinh \textit{et al.} \cite{7914660} both transfer the multiobjective optimization problem into a single-objective optimization problem by weighting coefficients. To maximize the total revenue, Wang \textit{et al.} \cite{7929399} formulate an optimization problem by jointly considering the offloading decision, resource allocation, and caching in heterogeneous wireless cellular networks and propose a distributed algorithm based on the alternating direction method of multipliers (ADMM). To address autonomous vehicles often have insufficient onboard resources to provide the required computation capacity,  Cui \textit{et al.} \cite{9112192} advocated a novel approach to offload computation-intensive autonomous driving services to roadside units and cloud. And combined an integer linear programming (ILP) formulation for offline optimization of the scheduling strategy and a fast heuristics algorithm for online adaptation. However, model-driven have issues such as long online processing time, which cannot meet the low latency requirements of connected and autonomous vehicles (CAV) tasks.

To address issues such as long online processing time for model-driven algorithms, some works adopt data-driven methods to manage the resource in task offloading. For instance, Dai  \textit{et al.} proposed a dynamic resource allocation architecture for computing and caching in \cite{dai2019artificial}, and utilized deep reinforcement learning (DRL) to maximize system utility. He \textit{et al.} used DRL to maximize a reward function defined by system utility in order to solve the joint optimization problem of communication, caching, and computation in \cite{he2017integrated}. Zhang \textit{et al.} \cite{zhang2022federated} proposed a distributed computing offloading algorithm to optimize task offloading latency. Zhang \textit{et al.}  \cite{8471165} utilized the cognitive radio (CR) to alleviate the spectrum scarcity problem during computation offloading. To reduce transmission costs among vehicle-to-infrastructure (V2I) and vehicle-to-vehicle (V2V) communications, they propose a deep Q-learning method to schedule the communication modes and resources. Li \textit{et al.}  \cite{9119487} studied a collaborative computing approach in vehicular networks and proposed a DRL technique to tackle a complex decision-making problem. Cheng \textit{et al.} \cite{8297294} proposed a novel DRL-based system with two-phase resource allocation and task scheduling to reduce energy costs for cloud service providers with large data centers and a large number of user requests with dependencies. Targeting the problem of multi-objective work-flow scheduling in the heterogeneous IaaS cloud environment. Wang \textit{et al.} \cite{8676306} modeled the multiobjective workflow scheduling as a stochastic Markov game and develop a decentralized multi-task deep reinforcement learning framework that is capable of obtaining correlated equilibrium solutions of workflow scheduling. 

In addition to the papers dedicated to the resource allocation of I-CC, the resource allocation methods of I-CSC are also emerging. Qi \textit{et al.} \cite{9439524} presented a traffic-aware task offloading mechanism that optimally combines communication and sensing abilities of serving nodes (SNs) to minimize overall response time (ORT) for computation-intensive and latency-sensitive vehicular applications, and used binary search algorithm to solve this problem. Gong \textit{et al.} \cite{9982429} proposed an environment-aware offloading mechanism (EAOM) based on an integrated computation, sensing and communication systems to minimize the ORT of the system and used deep optimization problem of deterministic policy gradient (DDPG).

The related work regarding I-CSC mainly concentrates on reducing the task latency subject to resource constraints. For CAV tasks, safety is the primary metric required to be guaranteed, often reflected in task latency. Ensuring the latency-sensitive tasks are completed within their maximum threshold is crucial for safety. Furthermore, it is worth pointing out that the latency reduction is at the expense of consuming more resources and energy, resulting in large system costs. In light of this, in this paper, we minimize the offloading cost consisting of energy consumption and resource payment under latency constraints.  To solve this problem, we propose a knowledge-driven algorithm, an innovative fusion of traditional algorithms with neural networks, to improve the performance.

\section{System Model}
In this section,  the vehicle-RSU cooperation architecture is proposed. Next, we provide a detailed exposition of the specific scenarios investigated, establish three computation modes and analyze the latency and cost of each mode. Finally, the total cost minimization problem is formulated with the communication and computing resources constraints in section 2.5.

\subsection{Offloading Framework with Integrated Sensing, Computation and Communication for Vehicular Networks}

\begin{figure}
\centering
\includegraphics[width=11cm]{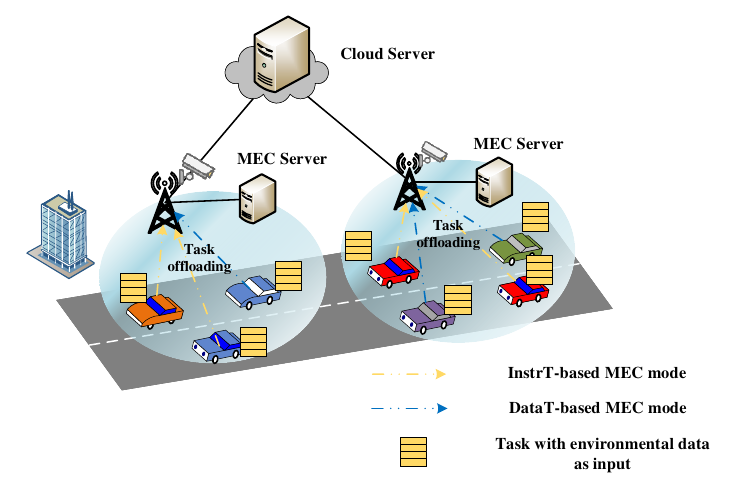} 
\caption{Vehicle-RSU cooperation architecture.}
\label{fig:1}
\end{figure}

As shown in Fig. \ref{fig:1}, the integrated sensing, computing and communication framework for vehicular networks consists of one centralized control center, several RSUs and multiple vehicles, where RSUs with MEC servers are deployed along the roadside to cover the partitioned road segments respectively.  Both vehicles and RSUs are equipped with an adequate number of sensors, such as high-definition cameras, LiDAR and millimeter-wave radar, to perceive the surrounding environmental information. The sensed real-time data serves as input for some computation-intensive and latency-intensive vehicular tasks, for instance, AR-based automotive driving with 3D Reconstruction. Due to the limited onboard computation resources in vehicles, such tasks are usually offloaded to RSUs with powerful computation ability.

For vehicles close to an RSU with multiple sensors, it is possible that the RSU can perceive similar environmental data as vehicles do. Although the perception data collected by the RSU and vehicles varies in perspective, such incongruities can be mitigated by providing coordinates of vehicles to the RSU to pre-conduct the coordinate transformation via matrix operation \cite{he2017integrated}. As a consequence, during the task offloading, each vehicle can select to transmit the simple task-related computation instruction with its coordinate, instead of massive environmental data, to the RSU, which remarkably decreases the volume of transmitted data. Armed with vehicles' coordinates and computation instructions, RSUs employ their own sensed environmental data to facilitate the computation offloading process. This novel I-CSC paradigm has the potential to significantly reduce the latency of computation-intensive vehicular tasks.

\begin{figure}
\centering
\includegraphics[width=15cm]{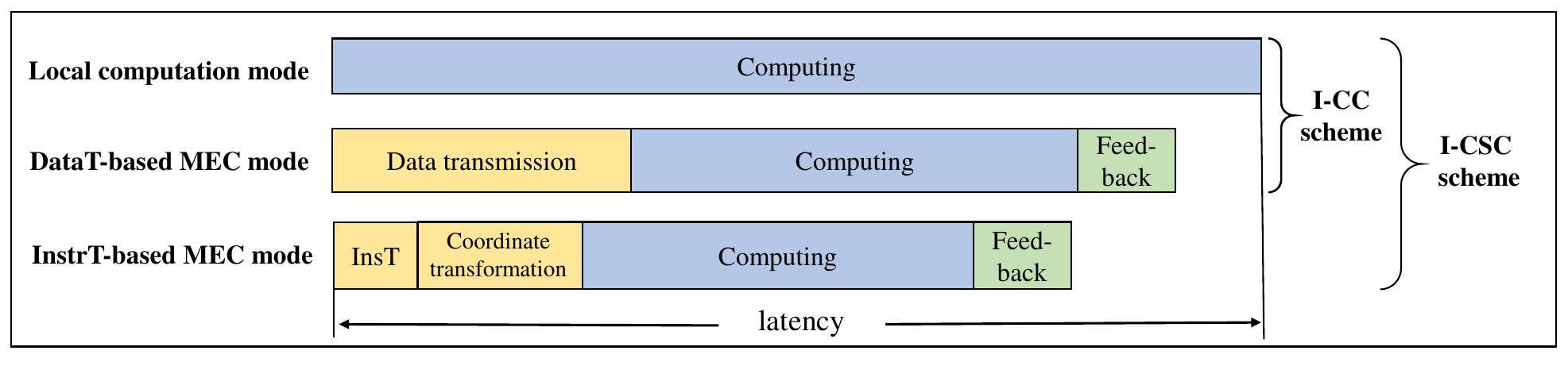} 
\caption{The latency of the considered three computation modes.}
\label{fig:2}
\end{figure}

In this paper, we consider the task offloading of vehicles with I-CSC in the coverage of one RSU, where the set of vehicles is defined by $\mathcal{I} = \{1, 2, ..., I\}$. Each vehicle carries a computation-intensive and latency-intensive task that takes environmental data as input. According to \cite{zhang2022federated}, such computing tasks can be arbitrarily split into two independent subtasks that run in parallel. Therefore, to improve the computation offloading efficiency, we consider a continuous task offloading strategy, where a portion of the task is locally processed by the vehicle and the remainder is simultaneously offloaded to the RSU for parallel computing. The task offloading ratio is denoted as $\eta_i \in [0,1]$, indicating the proportion of the task offloaded to the RSU. To keep the same with practical networks, in our paper, the orthogonal frequency division multiplexing (OFDM) communication technology is utilized, where each user occupies one subcarrier and there is no adjacent channel inference. As the RSU also perceives the environmental data, in the offloading process, vehicles can choose to transmit either the environmental data or the computation instruction to the RSU depending on the input data size and the network status, which is referred to as data transmission-based (DataT-based) or instruction transmission based (InstrT-based) MEC mode. The variable of transmission mode selection for task $i$, denoted as $\alpha_i$, is a binary value, where $\alpha_i = 1$ corresponds to the data transmission and $\alpha_i = 0$ indicates the computation instruction transmission. Therefore, as illustrated in Fig. \ref{fig:2}, the considered task offloading framework with I-CSC for vehicular networks consists of three computation modes, i.e., the local computation mode, the DataT-based MEC mode and the InstrT-based MEC mode. While the conventional task offloading is only with the integrated communication and computing, which includes the local computation mode and the DataT-based MEC mode.

Notice that latency is of particular significance for driving-related vehicular tasks and tasks generated by different vehicles in dynamic road situations may have various maximum latency tolerances. Together with the considered I-CSC  framework, the task for vehicle $i$ is defined as $\phi _i = \{t_i^{\text{max}}, c_i, b_i, S_i^{\text{instr}}, b_i^{\text{instr}}\}$, where $t_i^{\text{max}}$ signifies the maximum latency tolerance for the task, $c_i$ represents the task's input data size, $b_i$ denotes the required number of CPU cycles for the task computation, $S_i^{\text{instr}}$ indicates the data size of computation instruction and vehicles' coordinate, and $b_i^{\text{instr}}$ represents the required number of CPU cycles for the coordinate transformation. For each task, there are three possible computation modes in the considered I-CSC framework, whose latency compositions are respectively plotted in Fig. \ref{fig:2}. In addition to the latency, communication and computation resource cost is also an important metric that vehicle owners care about. In what follows, therefore, both the latency and the resource cost for each computation mode in the I-CSC framework are analyzed mathematically.


\subsection{Local Computation Mode}
When a portion of a task is processed locally by the onboard computation resource in the vehicle, its total latency is just the computing latency. Let $f_{i}^l$  represent the CPU frequency of vehicle $i$, the latency of task $i$ in the local computation mode is defined as 
\begin{equation}
    t_i^l=\frac{(1-\eta_i)b_i}{f^{l}_i}.
\end{equation}  

Similarly, the resource cost in the local computation mode only involves the computing energy cost. Let $e_2$ denote the cost of energy per unit consumed locally, then the energy cost of task $i$ in local computation mode is
\begin{equation}
    C_i^l=e_2(1-\eta_i)c_i(f_i^l)^2,
\end{equation}

\subsection{DataT-based MEC Mode}
For the DataT-based MEC mode, the input data of tasks sensed by the vehicle is transmitted to RSU and computed at the co-located MEC server. Hence, the latency of this mode includes the task uploading latency, the computing latency, and downlink feedback latency.

The latency in transmitting data from the vehicle to the RSU is
\begin{equation}
    t_{i}^{\text{data}}=\frac{\eta_{{i}}c_{i}}{W_{i}\log_2(1+\frac{P_{i}g_{i}}{\sigma^{2}})},
\end{equation}
where $W_i$ is the bandwidth allocated to the $i$-th vehicle user by RSU, $P_i$ is the transmit power of the i-th vehicle, $g_i$ represents the channel gain, and $\sigma^2$ is the noise power during the transmission process. Then, the computing latency of the RSU is
\begin{equation}
    t_i^{\text{comp}} = \frac{\eta_ib_i}{f_i},
\end{equation}
where $f_i$ is the CPU frequency allocated to task $i$ in the MEC server. As the amount of task result is too small, the downlink latency can be ignored. Therefore, the total latency in this mode is
\begin{equation}
    t_i^{D}=\frac{\eta_{i}c_i}{W_{i}\log_2(1+\frac{P_ig_i}{\sigma^2})}+\frac{\eta_\mathrm{i}b_i}{f_i}.
\end{equation}

During the process, the cost of completing the task is more complex than that of the local computation mode. It consists of two parts, one is the energy cost in the process of unloading transmission and computing. The other is the cost of bandwidth and computing resources to be paid to the RSU during resource allocation.

The energy cost in this mode is the energy consumed in the transmission process. The energy cost consumed in the transmission process of  vehicle  $i$ is as follows
\begin{equation}
    C_i^e=P_i\frac{(1-\alpha_i)\eta_ic_ie_1}{W_{i}\log_2(1+\frac{P_ig_i}{\sigma^2})},
\end{equation}
where $e_1$ represents the energy cost per unit of energy consumed during transmission.

The payment cost for vehicle $i$ to the RSU is expressed as
\begin{equation}
    C_i^d=\mu_1W_i+\mu_2f_i,
\end{equation}
where $\mu_1$  represents the payment cost per unit of bandwidth, and $\mu_2$ represents the payment cost per unit of computation resources.

The total cost in DataT-based MEC mode is
\begin{equation}
    C_i^D=P_i\frac{(1-\alpha_i)\eta_ic_ie_1}{W_{i}\log_2(1+\frac{P_ig_i}{\sigma^2})}+\mu_1W_i+\mu_2f_i.
\end{equation}

\subsection{InstrT-based MEC Mode}
When the sub-task is executed in RSU and the vehicle transmits the computation instruction to RSU, the latency includes the computation instruction transmission latency, instruction conversion latency, computation latency and downlink transmission latency.

The transmission latency when uploading computation instructions to RSU is
\begin{equation}
t_i^{\text{instr}}=\frac{\eta_{i}S_i^{\text{instr}}}{W_i \log_2(1+\frac{P_{i}g_i}{\sigma^2})}.
\end{equation}

To make the viewpoint of data obtained at the RSU consistent with the data that the vehicle wants to process at this time, it is necessary to preprocess the environmental data sensed by the RSU. This can be achieved by performing matrix operations using the vehicle coordinates included in the computing instruction. The conversion time is related to the size of the perceived data, which is expressed as
\begin{equation}
    t_i^{\text{tra}}=\frac{\eta_i b_i^{\text{intar}}}{f_i}.
\end{equation}

After the computing instruction arrives at RSU, the computing latency of the RSU is
\begin{equation}
    t_i^{\text{comp}} = \frac{\eta_ib_i}{f_i}.
\end{equation}

Since the transmission time of the computing instruction is small relative to the coordinate conversion time, we ignore the transmission time in this paper. Therefore, the total computation latency in InstrT-based MEC mode is
\begin{equation}
    t_i^{\text{In}} = \frac{\eta_i b_i^{\text{intar}}}{f_i}+\frac{\eta_ib_i}{f_i}.
\end{equation}

Compared with the DataT-based MEC mode, the energy consumption generated by InstrT-based MEC mode is particularly small and can be ignored. Hence, the total cost in InstrT-based MEC Mode is the payment cost paid by the vehicle $i$ to the RSU
\begin{equation}
    C_i^{\text{In}} = \mu_1W_i+\mu_2f_i.
\end{equation}
\subsection{Problem Formulation}
The latency and cost of the three modes are analyzed above. This paper stands from the perspective of the vehicle user. When the vehicle needs to process a task, the user hopes to complete the task with the least cost within the time requirement.

The latency of task completion depends on the longest latency of subtasks in the three modes. With the help of variables $\alpha_i$, the latency of DataT-based MEC mode and InstrT-based MEC mode can be merged, and the latency of task processing at RSU can be obtained as follows
\begin{equation}
    t_i^{\text{RSU}}=\alpha_i\frac{\eta_\mathrm{i}c_i}{W_{i}\log_2(1+\frac{P_ig_i}{\sigma^2})}+(1-\alpha_i)\frac{\eta_\mathrm{i}b_i^{\text{intar}}}{f_i}+\frac{\eta_\mathrm{i}b_i}{f_i}.
\end{equation}

The total latency to complete the task is the maximum value of the local processing latency and RSU processing latency. So the total latency $t_i$ is
\begin{equation}
    t_i = \text{max} \{t_i^{\text{RSU}},t_i^l \}.
\end{equation}

The cost of completing task $i$ is the sum of the respective costs under the three modes. So the cost of task $i$  is
\begin{equation}
C_i = P_i\frac{\alpha_i\eta_ic_ie_1}{W_{i}\log_2(1+\frac{P_ig_i}{\sigma^2})}
    + e_2(1-\eta_i)c_i(f_i^l)^2
    + \mu_1W_i+\mu_2f_i.    
\end{equation}

Let $G(\bm{\alpha,\eta,p,w,f})$ denote the total cost of the entire system, then
\begin{equation}
G(\bm{\alpha,\eta,p,w,f}) = \sum_{i=1}^{I}  C_i,    
\end{equation}
where  $\boldsymbol{\alpha} =[\alpha_1, \alpha_2, ..., \alpha_I ]^T$, $\boldsymbol{\eta} =[\eta_1, \eta_2, ..., \eta_I ]^T$, $\boldsymbol{p} =[P_1, P_2, ..., P_I ]^T$, $\boldsymbol{w} =[W_1, W_2, ..., W_I ]^T$ and  $\boldsymbol{f} =[f_1, f_2, ..., f_I ]^T$.


In this paper, we consider minimizing the total cost under the constraints of communication resources, computing resources, and latency conditions. We can optimize the transmission mode variable $\bm{\alpha}$, task partitioning variable $\bm{\eta}$, vehicle transmit power variable $\bm{p}$, and the base station's allocation of bandwidth and computation resources variables $\bm{w}$ and $\bm{f}$  to minimize the total cost. The final formulated problem is represented as 

\begin{subequations}
\begin{align}
\label{equ:6a}
\mathcal{P}1:
\underset{\; \bm{\alpha,\eta,p,w,f}}{\text{minimize}}&\qquad G(\bm{\alpha,\eta,p,w,f})\\
\label{equ:6b}
\text{s. t.} &\qquad  \text{C1}:\eta_i \in [0,1], \forall {i \in \{1,2,...,I\}}\\
\label{equ:6e}
&\qquad \text{C2}:\alpha_i = \{0,1\}, \forall {i \in \{1,2,...,I\}} \\
\label{equ:6c}
&\qquad \text{C3}:P_i \leq P_{\text{car}}, \forall {i \in \{1,2,...,I\}}\\
\label{equ:6d}
&\qquad \text{C4}:\sum_{i=1}^{I} f_i \leq f_{\text{RSU}}, \\
&\qquad \text{C5}:\sum_{i=1}^{I} W_i \leq W_{\text{RSU}},\\
&\qquad \text{C6}:t_i \leq t_i^{\text{max}}, \forall {i \in \{1,2,...,I\}},
\end{align}
\end{subequations}
where C1 represents the task partitioning ratio constraint, C2 represents the transmission mode constraint, C3 represents the maximum transmission power constraint for the vehicles with $P_{car}$ denoting the maximum transmit power of a vehicle, C4 represents the computation resource constraint for the BS with  $f_{RSU}$ being the total computation resource at the RSU, C5 represents the bandwidth constraint for the RSU with $W_{RSU}$ being the total bandwidth available at the RSU, and C6 represents the maximum latency constraint.
Due to $t_i = \text{max}\{t_i^{\text{RSU}}, t_i^{l}\}$, we can convert the $C6$ constraint into the following two constraints
\begin{equation}
    \text{C6a}:t_i^{\text{RSU}} \leq t_i^{\text{max}}, \forall {i \in \{1,2,...,I\}},
\end{equation}
\begin{equation}
    \text{C6b}:t_i^{l} \leq t_i^{\text{max}}, \forall {i \in \{1,2,...,I\}}.
\end{equation}

Then, the final objective optimization problem is equivalently expressed as
\begin{subequations}
\begin{align}
\label{equ:6a}
\mathcal{P}2:
\underset{\; \bm{\alpha,\eta,p,w,f}}{\text{minimize}}&\qquad G(\bm{\alpha,\eta,p,w,f})\\
\label{equ:6b}
\text{s. t.} &\qquad  \text{C1}:\eta_i \in [0,1], \forall {i \in \{1,2,...,I\}}\\
\label{equ:6e}
&\qquad \text{C2}:\alpha_i = \{0,1\}, \forall {i \in \{1,2,...,I\}} \\
\label{equ:6c}
&\qquad \text{C3}:P_i \leq P_{\text{car}}, \forall {i \in \{1,2,...,I\}}\\
\label{equ:6d}
&\qquad \text{C4}:\sum_{i=1}^{I} f_i \leq f_{\text{RSU}}, \\
&\qquad \text{C5}:\sum_{i=1}^{I} W_i \leq W_{\text{RSU}},\\
&\qquad \text{C6a}:t_i^{\text{RSU}} \leq t_i^{\text{max}}, \forall {i \in \{1,2,...,I\}},\\
&\qquad \text{C6b}:t_i^{l} \leq t_i^{\text{max}}, \forall {i \in \{1,2,...,I\}}.
\end{align}
\end{subequations}
Due to the coupling between variables, this problem is non-convex. 

\section{Model-based AM Algorithm for Task Offloading with I-CSC}

The original problem $\mathcal{P}2$ is highly challenging to solve directly due to its nonconvex nature \cite{boyd2004convex}, which is caused by the mutual coupling of $\boldsymbol{\alpha}, \boldsymbol{\eta}, \boldsymbol{p}, \boldsymbol{w}, \boldsymbol{f}$ in the objective function, as well as the maximum latency constraints. To tackle this in a computationally efficient manner, we employ the widely-used model-based AM algorithm, whose main idea is breaking down the original complex problem with multiple variables into sub-problems involving partial variables, solving them in turn \cite{9246287} while keeping the other variables fixed. In this section, we decompose problem $\mathcal{P}2$ into four sub-problems, i.e., transmission mode selection, task offloading ratio decision, transmission power allocation, as well as bandwidth and computing resource allocation, and alternatively tackle them to find a good solution.  

\subsection{Transmission Mode Selection}
We first solve the problem of transmission mode selection for each CAV. In the original problem $\mathcal{P}2$, the variable that determines the transmission mode of vehicle $i$ is the binary variable $\alpha_{i}$. When the other variables are fixed, the objective function of the transmission mode selection problem is defined as $g(\bm{\alpha})$, which is expressed as
\begin{subequations}
\begin{align}
\label{equ:_29a}
\mathcal{P}3: 
&\min_{\bm{\alpha}} \quad  \sum_{i=1}^{I} P_{i} \frac{\alpha_i \eta_{i} c_{i} e_{1}} {W _{i} \log_{2}(1+\frac{P_{i} g_{i}}{\sigma^{2}})} \\
\label{equ:_29b}
&\text{s. t.} \qquad  \alpha_{i}\in \{0,1\}, \forall {i \in \{1,2,...,I\}},\\
\label{equ:_29c}
&\qquad \quad  \frac{\alpha_{i} \eta_{i} c_{i}}{W_{i} \log_{2}(1+\frac{P_i g_i}{\sigma^2})}+(1-\alpha_{i})\frac{\eta_{i} b_{i}^\text{intar}}{f_i}+\frac{\eta_{i} b_{i}}{f_{i}} \leq t_{i}^{\text{max}} , \forall {i \in \{1,2,...,I\}}.
\end{align}
\end{subequations}

    Observing from problem $\mathcal{P}$3, one can find that the objective function is a linear combination of the decision variable $\alpha_i$, and there is no coupling of the transmission choices between different vehicles in the constraints. Therefore,  problem $\mathcal{P}$3 can be divided into $I$ parallel problems with each aiming to optimize a single variable $\alpha_i$. Specifically, the problem of optimizing $\alpha_i$ is expressed as
    \begin{subequations}
    \begin{align}
    \label{equ:_29a}
    \mathcal{P}3': 
    &\min_{\alpha_i} \quad P_{i} \frac{\alpha_i \eta_{i} c_{i} e_{1}} {W _{i} \log_{2}(1+\frac{P_{i} g_{i}}{\sigma^{2}})} \\
    \label{equ:_29b}
    &\text{s. t.} \qquad  \alpha_{i}\in \{0,1\}, \\
    \label{equ:_29c}
    &\qquad \quad  \frac{\alpha_{i} \eta_{i} c_{i}}{W_{i} \log_{2}(1+\frac{P_i g_i}{\sigma^2})}+(1-\alpha_{i})\frac{\eta_{i} b_{i}^\text{intar}}{f_i}+\frac{\eta_{i} b_{i}}{f_{i}} \leq t_{i}^{\text{max}}.
    \end{align}
    \end{subequations}
    
Note that $\mathcal{P}3'$ is a binary linear programming problem with constraints, which is easy to optimally solve by the implicit enumeration method.

\subsection{Task Offloading Ratio Decision}
The second sub-problem we face is the task offloading ratio decision, where we fix the variables $ \boldsymbol{p},\boldsymbol{w},\boldsymbol{f} $ and $\boldsymbol{\alpha}$ to obtain the optimal value of $\boldsymbol{\eta}$ for that case. The constraints considered are C1, C6a, and C6b. The objective function is defined as $g(\bm{\eta})$, and represented as follows

\begin{subequations}
\begin{align}
\label{equ:_30a}
\mathcal{P}4: 
& \min_{\bm{\eta}} \quad \sum_{i=1}^{I}(P_{i} \frac{\alpha_i \eta_{i} c_{i} e_{1}} {W _{i} \log_{2}(1+\frac{P_{i}g_{i}}{\sigma^{2}})}+e_{2}(1-\eta_{i})c_{i}{f^{l}_i}^{2}+\mu_{1} W_{i}+\mu_{2}f_{i})\\
\label{equ:_30b}
&\text{s. t.} \quad \eta_{i} \in[0,1],  \forall {i \in \{1,2,...,I\}},\\
& \qquad ~~ \eta_{i} \leq \frac{t_{i}^{\max }}{\alpha_{i} {\frac{c_{i}}{W_{i} \log _{2}\left(1+\frac{P_{i} g_{i}}{\sigma^{2}}\right)}}+(1-\alpha_{i}) \frac{b_{i}^{\text{intar}}}{f_{i}}+\frac{b_{i}}{f_{i}}}, \forall {i \in \{1,2,...,I\}},\\
& \qquad ~~ \eta_{i} \leq 1-\frac{t_{i}^{\max } f^{l}_i}{b_{i}}, \forall {i \in \{1,2,...,I\}}.
\end{align}
\end{subequations}

Similar to the first sub-problem, the decision $ \eta_{i}$ under different tasks has no coupling effect in the objective function and constraints, so we can still decompose the original problem into sub-problems in the single-vehicle case. After eliminating the terms in the optimization objective function that are not related to $\eta_{i}$, the sub-problem can be simplified as 

\begin{align}
\mathcal{P}4':
& \underset{\eta_i}{\operatorname{minimize}}\left(P_{i} \frac{\alpha_{i} c_{i} e_{1}}{W_{i} \log _{2}\left(1+\frac{P_{i} g_{i}}{\sigma^{2}}\right)}-e_{2} c_{i}\left(f^{l}_i\right)^{2}\right) \eta_i \\
& \text { s. t. } \quad\eta_i \in[0,1],\\
& \qquad \quad  \eta_i \leq \frac{t_{i}^{\max }}{\alpha_{i} \frac{c_{i}}{W_{i} \log _{2}\left(1+\frac{P_{i} g_{i}}{\sigma^{2}}\right)} + (1-\alpha_{i}) \frac{b_{i}^\text{intar}}{f_{i}}+\frac{b_{i}}{f_{i}}}, \\
&\qquad \quad  \eta_{i} \leq 1-\frac{t_{i}^{\max }f^{l}_i}{b_{i}}.
\end{align}
Then, the three constraints in problem $\mathcal{P}4'$ can be equivalently transformed into the following form
\begin{align}
\label{sub2-c1}
\eta_i \in\left[\max \left(0,1-\frac{t_{i}^{\max } f^{l}_i}{b_{i}}\right), \min \left(1, \frac{t_{i}^{\max }}{\alpha_{i} \frac{c_{i}}{W_{i} \log _{2}\left(1+\frac{P_{i} g_{i}}{\sigma^{2}}\right)}+\left(1-\alpha_{i}\right) \frac{b_{i}^\text{intar}}{f_{i}}+\frac{b_{i}}{f_{i}}}\right)\right] 
\end{align}
with
\begin{align}
\label{sub2-c2}
 1-\frac{t_{i}^{\max } f^{l}_i}{b_{i}} \leq \min \left(1, \frac{t_{i}^{\max }}{\alpha_{i} \frac{c_{i}}{W_{i} \log _{2}\left(1+\frac{P_{i} g_{i}}{\sigma^{2}}\right)}+\left(1-\alpha_{i}\right) \frac{b_{i}^\text{intar}}{f_{i}}+\frac{b_{i}}{f_{i}}}\right), 
\end{align}
where (\ref{sub2-c2}) is to ensure that problem $\mathcal{P}4'$ is feasible.

We observe that the problem is a simple single-variable linear optimization problem with constraints, and the optimal solution can be obtained by Newton's method. 

\subsection{Transmission Power Allocation} 

In the transmission power allocation problem, there is also no coupling between the transmission powers of different vehicles, so the power allocation problem can still be decomposed into a sub-problem for the single-vehicle case, where the constraints that play a role are C3 and C6a. The objective function is defined as $g(\bm{p})$, and expressed as follows

\begin{subequations}
\begin{align}
\label{equ:__31a}
\mathcal{P}5: 
&\min_{P_i} \quad  P_{i} \frac{\alpha_i \eta_{i} c_{i} e_{1}} {W _{i} \log_{2}(1+\frac{P_{i} g_{i}}{\sigma^{2}})}\\
\label{equ:__31b}
&\text{s. t.} \quad ~~ P_{i} \leq P_{\text{car}}, \\
\label{equ:__31c}
&\qquad \quad P_{i} \geq \frac{\sigma^2(e^{\frac{\eta_{i}c_{i}}{W_{i}(t_{i}^{\max} - (1-\alpha_{i})\frac{\eta_{i} b_{i}^\text{intar}}{f_i}+\frac{\eta_{i} b_{i}}{f_{i}})}}-1)}{g_{i}}.
\end{align}
\end{subequations}

The objective function in this problem is monotonically increasing, and the proof procedure is as follows
\begin{align}
    \frac{\partial \frac{ P_{i} \alpha_i \eta_{i} c_{i} e_{1}} {W _{i} \log_{2}(1+\frac{P_{i} g_{i}}{\sigma^{2}})}}{\partial P_{i}} = \frac{\alpha_i \eta_{i} c_{i} e_{1}} {W _{i} \log_{2}(1+\frac{P_{i} g_{i}}{\sigma^{2}})} - \frac{ P_{i} \alpha_i \eta_{i} c_{i} e_{1}W_{i}  \frac{g_{i}}{\sigma^{2}}  \ln 2 } {(W _{i} \log_{2}(1+\frac{P_{i} g_{i}}{\sigma^{2}}))^2(1+\frac{P_{i} g_{i}}{\sigma^{2}})}   > 0.
\end{align}
Thus, when the feasible domain is nonempty, the optimal power can be obtained as 
\begin{equation}
\label{equ:33power}
    P_{i} = \frac{\sigma^2(e^{\frac{\eta_{i}c_{i}}{W_{i}(t_{i}^{max} - (1-\alpha_{i})\frac{\eta_{i} b_{i}^\text{intar}}{f_i}+\frac{\eta_{i} b_{i}}{f_{i}})}}-1)}{g_{i}}.
\end{equation}

\subsection{Bandwidth and Computing Resource Allocation}

Finally, we solve the allocation of bandwidth and computational resources, in which the variables are bandwidth $\boldsymbol{w}$ and computational resources $\boldsymbol{f}$. The constraints associated with this are C4, C5, and C6a, and the objective function is defined as $g(\bm{w, f})$. Mathematically, the bandwidth and computing resource allocation problem is expressed as follows 
\begin{subequations}
\begin{align}
\label{equ:_32a}
\mathcal{P}6: 
&\min_{\bm{w,f}} \quad \sum_{i=1}^{I} P_{i} \frac{\alpha_i \eta_{i} c_{i} e_{1}} {W _{i} \log_{2}(1+\frac{P_{i} g_{i}}{\sigma^{2}})} + e_{2}(1-\eta_{i})c_{i}{f^{l}_i}^{2}+\mu_{1} W_{i} +\mu_{2}f_{i} \\
\label{equ:_32b}
&\text{s. t.} \qquad \sum_{i=1}^{I}f_{i}\leq f_{\text{RSU}},\\
\label{equ:_32c}
& ~~\quad\qquad \sum_{i=1}^{I} W_i \leq W_{\text{RSU}} ,\\
\label{equ:_32d}
&\qquad \quad \frac{\alpha_{i} \eta_{i} c_{i}}{W_{i} \log_{2}(1+\frac{P_i g_i}{\sigma^2})}+(1-\alpha_{i})\frac{\eta_{i} b_{i}^\text{intar}}{f_i}+\frac{\eta_{i} b_{i}}{f_{i}} \leq t_{i}^{\max}, \forall {i \in \{1,2,...,I\}}.
\end{align} 
\end{subequations}
    As shown in problem $\mathcal{P}$6, the objective function is a non-negative linear weighted summation of functions $\frac{1}{W_i}$ and $f_i$, which are respectively related to decision variables $W_i$ and $f_i$. Notice that the fractional function $\frac{1}{W_i}$ with $W_i>0$ is convex and linear function $f_i$ is also convex. According to the convexity preservation theorem that the non-negative linear weighted summation of convex functions is also convex \cite{boyd2004convex}, the objective function in $\mathcal{P}$6 is a convex function. In a similar way, one can also prove that all constraints in problem $\mathcal{P}$6 are convex. As a consequence, problem $\mathcal{P}$6 is a convex problem.  Therefore, it can be optimally solved using the Lagrangian dual decomposition method.
    
    Let $u$, $v$, and $\bm{q}$ be Lagrangian multipliers with $\bm{q} = \{q_1, q_2, ..., q_I \}$,  the Lagrangian function is expressed as 
\begin{equation}
    \mathcal{L}(\bm{w},\bm{f},u,v,\bm{q}) = G(\bm{w}, \bm{f}) + u(\sum_{i=1}^{I}f_{i} - f_{\text{RSU}})+v(\sum_{i=1}^{I} W_i - W_{\text{RSU}})+ \sum_{i=1}^{I} q_i (T_{i}(W_i, f_i)),
\end{equation}
where
\begin{align}
    G(\bm{w}, \bm{f}) = \sum_{i=1}^{I}(P_{i} \frac{\alpha_i \eta_{i} c_{i} e_{1}} {W _{i} \log_{2}(1+\frac{P_{i} g_{i}}{\sigma^{2}})} + e_{2}(1-\eta_{i})c_{i}{f^{l}_i}^{2}+\mu_{1} W_{i} +\mu_{2}f_{i}),\\
    T_{i}(W_i, f_i) = ( \frac{\alpha_{i} \eta_{i} c_{i}}{W_{i} \log_{2}(1+\frac{P_i g_i}{\sigma^2})}+(1-\alpha_{i})\frac{\eta_{i} b_{i}^\text{intar}}{f_i}+\frac{\eta_{i} b_{i}}{f_{i}} - t_{i}^{\max}).
\end{align}
As a result, the dual function of the original problem $\mathcal{P}$6 is expressed as
\begin{equation}
    h(u,v,\bm{q}) = \min_{\bm{w},\bm{f}} \mathcal{L}(\bm{w},\bm{f},u,v,\bm{q})
\end{equation}
and the dual problem is represented as
\begin{equation}
   \max_{u,v,\bm{q}} h(u,v,\bm{q}) =\max_{u,v,\bm{q}} \min_{\bm{w},\bm{f}} \mathcal{L}(\bm{w},\bm{f},u,v,\bm{q})
\end{equation}
As directly obtaining the closed-form solution of the dual problem is difficult, an iterative algorithm is adopted to tackle this problem by alternatively updating primal variables $\{\bm{w}, \bm{f}\}$  and Lagrangian multipliers $\{u, v, \bm{q}\}$.

With the given Lagrangian multiples, the optimal bandwidth resource and computational resource allocation with closed forms in the $n$-th iteration can be derived as 

\begin{align}
&\quad\frac{\partial\mathcal{L}(\bm{w},\bm{f},u,v,\bm{q})}{\partial \bm{w}} = 0 ,
\label{bindwidth}
&W_{i}^n=\sqrt{\frac{P_i\alpha_i\eta_i c_i e_1 + q_i\alpha_i\eta_i c_i}{(\mu_1+v_i)\log_2(1+\frac{P_i g_i}{\sigma^2})}},\\
&\quad\frac{\partial\mathcal{L}(\bm{w},\bm{f} ,u,v,\bm{q})}{\partial \bm{f}}=0,
\label{computing}
&f_{i}^n=\sqrt{\frac{q_i[(1-\alpha_i)\eta_i c_{i,j}b_{i}^\text{intar}+\eta_i b_i]}{(\mu_2 + u_i)}}.
\end{align}

Then, Lagrange multipliers in the $n$-th iteration are iteratively updated by 
\begin{align}
\label{u_up}
& u_{}^{n+1} = [u_{}^{n} + \lambda_{1}(\sum_{i=1}^{I}f_{i} - f_{RSU}) ]^+, \\
\label{v_up}
& v_{}^{n+1} = [v_{}^{n} + \lambda_{1}(\sum_{i=1}^{I} W_i - W_{RSU})]^+ ,\\
\label{q_up}
& q_{i}^{n+1} = [q_{i}^{n} + \lambda_{2}(\frac{\alpha_{i} \eta_{i} c_{i}}{W_{i} \log_{2}(1+\frac{P_i g_i}{\sigma^2})}+(1-\alpha_{i})\frac{\eta_{i} b_{i}^\text{intar}}{f_i}+\frac{\eta_{i} b_{i}}{f_{i}} - t_{i}^\text{max})]^+ ,
\end{align}
where $\lambda_1,\lambda_2$ are step sizes. Due to the different update orders of magnitude, two different positive step sizes are used. 

The overarching algorithmic procedure for this subproblem unfolds as follows. We begin by establishing the iteration count for variable updates. Each iteration for updating variables $\bm{w}$ and $\bm{f}$ with the given Lagrangian multipliers. Then, Lagrangian multipliers are updated with given  $\bm{w}$ and $\bm{f}$. This cycle of updates for variables $\bm{w}$, $\bm{f}$ and the Lagrangian multipliers perpetuates until the designated number of iterations is reached. Upon completion, the final values of $\bm{w}$ and $\bm{f}$ are optimized solutions.

\subsection{Overall Alternating Minimization Algorithm for Computation Offloading}

\begin{algorithm}[h]
\caption{ AM Algorithm}
\begin{algorithmic}[1]
    \STATE Input: random initialization ${\boldsymbol{\alpha}}^0, {\boldsymbol{\eta}}^0$, ${\boldsymbol{p}}^0$, ${\boldsymbol{w}}^0$, ${\boldsymbol{f}}^0$, number of outer loops $K$;
    \FOR{$k\gets 0$ to $K$}
    \STATE Calculate transmission mode selection ${\boldsymbol{\alpha}}^k$ and obtain the solution to problem $\mathcal{P}3$ by utilizing one-dimensional search;
    \STATE Calculate task Offloading ratio decision ${\boldsymbol{\eta}}^k$ and obtain the solution to problem $\mathcal{P}4$ by utilizing one-dimensional search;
    \STATE Calculate transmission power ${\boldsymbol{p}}^k$ and obtain the solution to problem $\mathcal{P}5$ based on (\ref{equ:33power});
    \STATE Initialize  Lagrangian multiplier $u^0, v^0, \bm{q}^0$;
   \FOR{$n\gets 0$ to $N$}
    \STATE  Calculate bandwidth and computing resource  $\bm{w}^n$, $\bm{f}^n$ based on (\ref{bindwidth}), (\ref{computing});
    \STATE Update Lagrangian multiplier $u^n, v^n, \bm{q}^n$ according to (\ref{u_up}), (\ref{v_up}), (\ref{q_up});
    \ENDFOR
    \STATE $\bm{w}^k=\bm{w}^N$, $\bm{f}^k=\bm{f}^N$;
    \ENDFOR
    \end{algorithmic}
\end{algorithm}

\begin{figure}
    \centering
    \includegraphics[width=15cm]{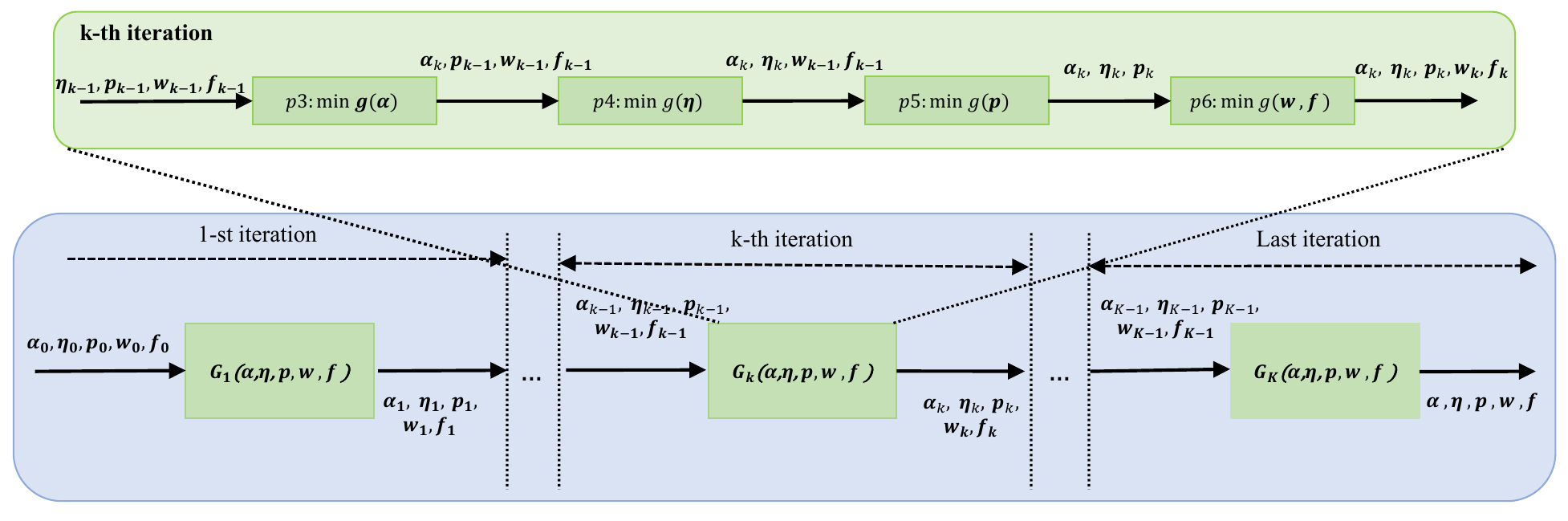} 
    \caption{The relationship of four subproblems in the AM algorithm.}
    \label{fig:re}
    \end{figure}

In this subsection, the overall AM algorithm of problem $\mathcal{P}$2 is summarised, which is outlined in Algorithm 1. The AM Algorithm initiates by setting random initial values to determine specific parameters and the aggregate number of iterations as $K$.  The algorithm systematically undertakes updates across four sub-problems. Specifically, during the $k$-th iteration,  the transmission mode selection is updated first, considering other variables as fixed inputs. Following this, the algorithm leverages a one-dimensional search approach to address problem $\mathcal{P}3$, updated the transmission mode selection  ${\boldsymbol{\alpha}}^k$. Progressing forward, using the updated ${\boldsymbol{\alpha}}^k$ obtained from the $k$-th iterations and  ${\boldsymbol{\eta}}^{k-1}$, ${\boldsymbol{p}}^{k-1}$, $\bm{w}^{k-1}$, $\bm{f}^{k-1}$ obtained from the $k-1$ th iterations 
 as fixed inputs,  problem $\mathcal{P}4$ is solved to determine the task offloading ratio ${\boldsymbol{\eta}}^k$. Subsequently, the transmission power allocation  ${\boldsymbol{p}}^k$ is updated by handling problem $\mathcal{P}5$. The final sub-problem $\mathcal{P}6$ is solved by the Lagrangian dual decomposition technique, which is an iterative algorithm updating primal variables $\{\bm{w}, \bm{f}\}$  and Lagrangian multipliers $\{u, v, \bm{q}\}$ alternatively.  Upon finalizing the $k$-th iteration, all pertinent variables are updated to their latest states, poised for the subsequent iteration.    After fulfilling the designated $K$ iterations, the algorithm yields the updated variables as the final resource allocation strategy of the AM algorithm. The relationship of the four subproblems in the AM algorithm is illustrated in Fig. \ref{fig:re}.

For the computational complexity analysis,  we observe that Algorithm 1 includes an outer loop with $K$ iterations. In each outer loop, four subproblems are alternatively tackled. As the calculation of the first three subproblems is relatively simple, the complexity of each outer loop is dominated by the fourth subproblem, which is an iterative algorithm with $N$ inner loops. In each inner loop,  the primal variables and dual variables need to be updated $3I+2$ times. Hence, the computational complexity of Algorithm 1 is $\mathcal{O}(KN(3I+2))$, which is simplified as $\mathcal{O}(KNI)$.

\section{Structural Knowledge-Driven Meta-Learning for I-CSC-based Task Offloading}
Although problem $\mathcal{P}$2 can be tackled by the model-based AM algorithm with explainability, the obtained solution is usually locally optimal for the overall problem even if each sub-problem achieves its global solution. This is because, in the AM algorithm, the sub-problem is optimized to minimize the local objective function other than the global objective function. Furthermore, the high computational complexity of the AM algorithm leads to long online processing time, which fails to keep pace with the rapid response requirements of driving-related tasks. As with powerful approximation ability and fast online inference, the machine learning method has attracted lots of attention in wireless resource allocation problems. However, the ``black-box" nature and weak interpretability hinder its widespread application in wireless networks. To tackle these challenges above, in this paper, we propose a novel structural knowledge-driven meta-learning method involving both the explainable AM algorithm and the neural network to handle the non-convex problem  $\mathcal{P}2$. In the following, the framework of the proposed SKDML is first introduced and the specific SKDML method for the I-CSC-based task offloading problem is then presented in detail. 

\subsection{The Proposed Structural Knowledge-Driven Meta-Learning Framework}
To simultaneously exploit the interpretability of model-based AM algorithms and the strong approximation ability of neural networks, in this paper, a novel SKDML method combining AM algorithm and neural network is proposed to solve the non-convex I-CSC-based task offloading problem $\mathcal{P}$2.  
As illustrated in Fig. \ref{fig:3}, the proposed SKDML framework, motivated by \cite{xia2022metalearning}, maintains the original inner-and-outer iterative structure of the model-base AM algorithm, referred to as structural knowledge in this paper, where four sub-problems are alternatively handled in the outer loop and each sub-problem is iteratively optimized in the inner loop. Based on this framework inspired by structural knowledge, the handcrafted iterative algorithmic principle for each sub-problem in the inner loop is replaced by an adaptive neural network to dynamically mimic the gradient-descent-based iterative process, forming a customized optimizer. According to the optimization theory \cite{NEURIPS2020_3d2d8ccb}, not only the current gradient but also the previous gradients have an impact on the variable update in an optimization problem. Hence, the recurrent neural network with the capability of memorizing past information, particularly the LSTM, is employed in the optimizer. As LSTM in the proposed SKDML framework is to learn a good learning optimizer for new tasks, instead of learning a neural network to directly map the input and output of tasks, the proposed framework belongs to the category of meta-learning. 

Furthermore, to pull out the solution from the local optimum, the proposed SKDML framework adopts the global objective function in problem $\mathcal{P}$2 as the global loss function to update the inner LSTM in the outer loop. Specifically, in the inner loop, the parameters of LSTM are frozen and variables of sub-problems are iteratively updated by the given LSTM-based optimizer as traditional gradient-descent-based algorithms do. After several inner iterations, the LSTM parameters are updated in the outer loop to minimize the global loss function, where widespread built-in optimizers like Adam are usually applied in this process. As a consequence, global loss information in the outer loop can be learned by the solution of each sub-problem in the inner loop to achieve superior performance, which is significantly different from the model-based AM algorithm that optimizes sub-problems with local objective functions. Moreover, the proposed SKDML framework is able to train in an unsupervised manner, suitable for solving non-convex optimization problems with no labeled data.



\subsection{The Proposed SKDML for I-CSC-based Task Offloading Problem}
%



\begin{figure}[htp]
    \centering
    \includegraphics[width=17cm]{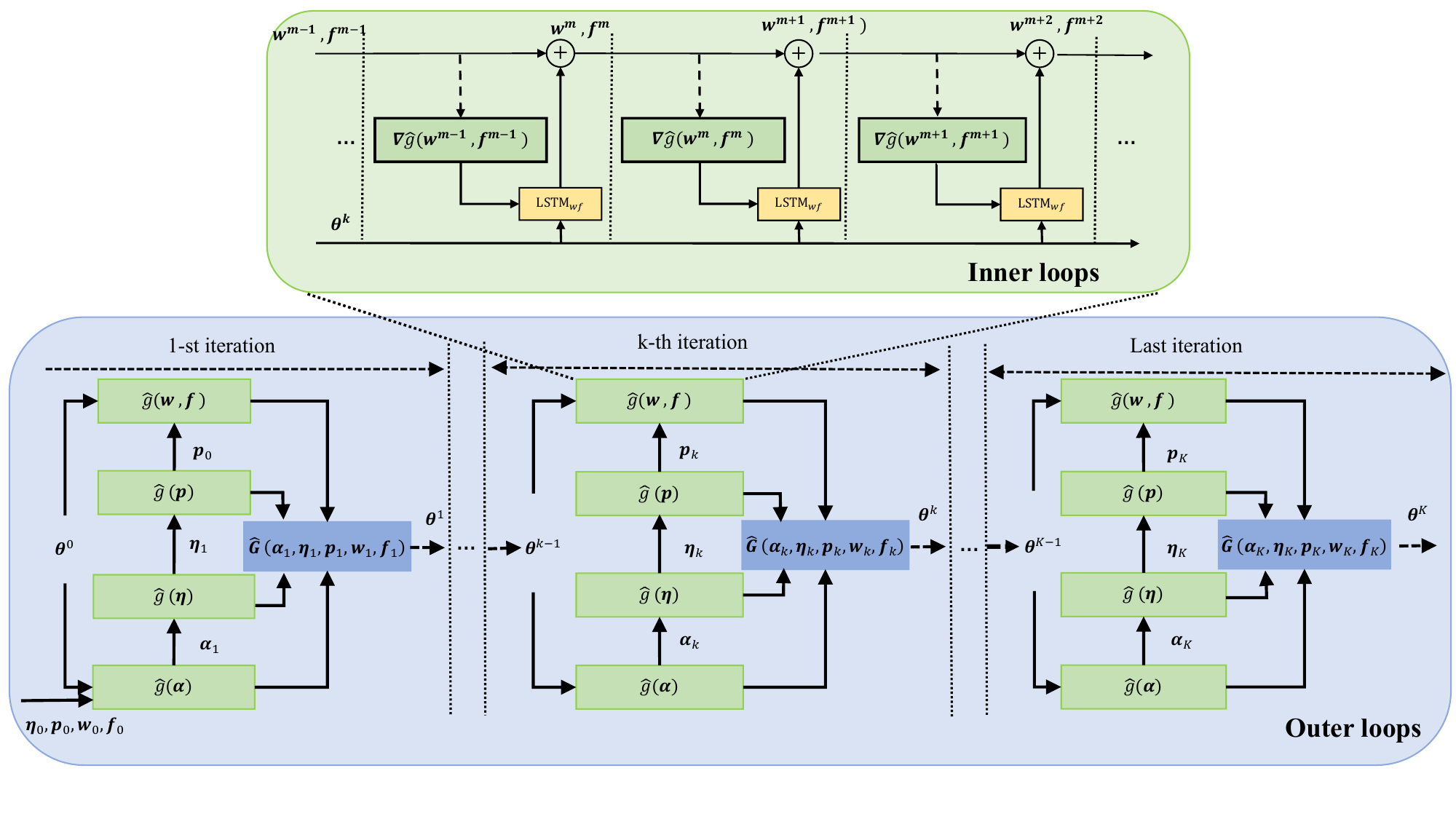}
    \caption{ The overall structure of SKDML algorithm. The algorithm architecture is divided into inner and outer loops. In the inner loop (blue box), the four LSTMs update the variables of the subproblem separately without updating the network parameters. In the outer loop (green box), the network parameters of the four LSTMs are updated using global loss. }
    \label{fig:3}
\end{figure}

This subsection elaborates on the proposed SKDML method for the non-convex I-CSC-based task offloading problem $\mathcal{P}$2. As shown in Fig. \ref{fig:3}, the main framework is guided by the alternative structure of the AM algorithm, consisting of an inner loop to iteratively update variables of the sub-problem and an outer loop to alternatively optimize four different sub-problems, i.e., the transmission mode selection problem, the task offloading radio decision problem, the transmission power allocation problem, the bandwidth and computing resource allocation problem. To solve problem $\mathcal{P}$2 with the proposed SKDML method, we add the constraint C6 as a penalty term to the objective function and reformulate the problem as  
\begin{equation}
\min \hat{G}({\boldsymbol{\alpha}}, {\boldsymbol{\eta}}, {\boldsymbol{p}}, {\boldsymbol{w}}, {\boldsymbol{f}})=G({\boldsymbol{\alpha}}, {\boldsymbol{\eta}}, {\boldsymbol{p}}, {\boldsymbol{w}}, {\boldsymbol{f}})+ \sum_{i=1}^{I} q_i(t_i-t_i^\text{max}),\quad \text{s. t.} \quad C1-C5,
\end{equation}
where $q_i$ represents for latency penalty factor.  We refer to the minimization of $\hat{G}({\boldsymbol{\alpha}}, {\boldsymbol{\eta}}, {\boldsymbol{p}}, {\boldsymbol{w}}, {\boldsymbol{f}})$ as the overall problem while to the minimization of $\hat{g}(\boldsymbol{\alpha})$, $\hat{g}(\boldsymbol{\eta})$, $\hat{g}(\boldsymbol{p})$ and $\hat{g}({\boldsymbol{w}}, {\boldsymbol{f}})$ as the four sub-problems to be optimized sequentially in each iteration.
In the following, we present both the inner variable update process for each sub-problem and the outer sub-problems alternating in detail.  


\subsubsection{The Variable Updating of Each Sub-Problem in the Inner Loop} 
In the inner loop, different from the traditional handcrafted gradient descent algorithm, variables in each sub-problem are iteratively updated by an adaptive and customized LSTM-based optimizer, whose parameters denoted by $\boldsymbol{\theta}$ are frozen. The reason for choosing LSTM in the proposed SKDML method is its ability to record the previous variable’s information by the cell states $\boldsymbol{C}$, which can be adaptively adjusted by the control gates inside the LSTM.   Inspired by the gradient descent algorithm, the LSTM-based optimizer also takes the gradient of the objective function, represented by $\nabla \hat{g}()$, and the accumulated previous state of the variable, represented by the cell states of LSTM $\boldsymbol{C}$, as inputs.  The outputs of the LSTM optimizer are the updating interval of the variable as well as the updated cell state of LSTM. 

Specifically, the first sub-problem is the transmission mode selection problem $\mathcal{P}3$, i.e., which is the minimization of $\hat{g}(\boldsymbol{\alpha})$. The LSTM network built for updating $\boldsymbol{\alpha}$, the parameters of the corresponding LSTM network, and the state cell are denoted as  $\text{LSTM}_{\boldsymbol{\alpha}}$, $\boldsymbol{\theta}_{\boldsymbol{\alpha}}$ and $\boldsymbol{C}_{\boldsymbol{\alpha}}$, and respectively. The variable $\boldsymbol{\alpha}$ updates using $\text{LSTM}_{\boldsymbol{\alpha}}$ are represented as
\begin{align}
\label{n_up}
    &\boldsymbol{\alpha}^n=\boldsymbol{\alpha}^{n-1}+\text{LSTM}_{\boldsymbol{\alpha}}\left(\nabla \hat{g}(\boldsymbol{\alpha}^{n-1}),\boldsymbol{C}_{\boldsymbol{\alpha}^{n-1}}, \boldsymbol{\theta}_{\boldsymbol{\alpha}}\right),
\end{align}
where $n=1,...,N$ is the $n$-th iteration for updating $\boldsymbol{\alpha}$ in the inner loop. To strictly satisfy the constraints C2 in problem $\mathcal{P}$3, we control $\boldsymbol{\alpha}$ within the range of $[0,1] $ and then use rounding to obtain the result.


The second sub-problem is the task offloading ratio decision problem $\mathcal{P}$4, i.e., the minimization of $\hat{g}(\boldsymbol{\eta})$. Denote by $\text{LSTM}_{\boldsymbol{\eta}}$, $\boldsymbol{\theta}_{\boldsymbol{\eta}}$ and $\boldsymbol{C}_{\boldsymbol{\eta}}$ the LSTM network established for updating $\boldsymbol{\eta}$, the parameters and the state cells of the corresponding LSTM network, respectively. The update of variable $\boldsymbol{\eta}$ by $\text{LSTM}_{\boldsymbol{\eta}}$-based optimizer in the inner loop is expressed as

\begin{align}
\label{j_up}
    &\boldsymbol{\eta}^j=\boldsymbol{\eta}^{j-1}+\text{LSTM}_{\boldsymbol{\eta}}\left(\nabla \hat{g}(\boldsymbol{\eta}^{j-1}),\boldsymbol{C}_{\boldsymbol{\eta}^{j-1}}, \boldsymbol{\theta}_{\boldsymbol{\eta}}\right),
\end{align}
where $j=1,...,J$ is the $j$-th iteration for updating $\boldsymbol{\eta}$ in the inner loop. To strictly satisfy the constraints C1 in problem $\mathcal{P}$4, we project the constraint to the range [0, 1]. 

The third sub-problem is the transmission power allocation, termed $\mathcal{P}$5, aiming at the minimization of $\hat{g}(\boldsymbol{p})$. The LSTM network constructed for the update of $ \boldsymbol{p}$is denoted as $\text{LSTM}_{\boldsymbol{p}}$, with its parameters and state cells represented by $\boldsymbol{\theta}_{\boldsymbol{p}}$ and $\boldsymbol{C}_{\boldsymbol{p}}$, respectively. The update to the variable  $ \boldsymbol{p}$ through the $\text{LSTM}_{\boldsymbol{p}}$ -based optimizer in the inner loop is articulated as
\begin{align}
\label{m_up}
    &\boldsymbol{p}^m=\boldsymbol{p}^{m-1}+\text{LSTM}_{\boldsymbol{p}}\left(\nabla \hat{g}(\boldsymbol{p}^{m-1}),\boldsymbol{C}_{\boldsymbol{p}^{m-1}}, \boldsymbol{\theta}_{\boldsymbol{p}}\right),
\end{align}
where $m=1,...,M$ is the $m$-th iteration for updating $\boldsymbol{p}$ in the inner loop. To strictly satisfy the constraints C3 in problem $\mathcal{P}$5, we map $\boldsymbol{p}$ into constraints C3.

The fourth sub-problem is the bandwidth and computational resource allocation, termed $\mathcal{P}$6, aiming at the minimization of $\hat{g}(\boldsymbol{w})$ and $\hat{g}(\boldsymbol{f})$. The LSTM network constructed for the update of $\hat{g}(\boldsymbol{w})$ and $\hat{g}(\boldsymbol{f})$ is denoted as $\text{LSTM}_{\boldsymbol{wf}}$, with its parameters and state cells represented by $\boldsymbol{\theta}_{\boldsymbol{wf}}$ and $\boldsymbol{C}_{\boldsymbol{wf}}$, respectively. The update to the variable  $ \boldsymbol{wf}$ through the $\text{LSTM}_{\boldsymbol{wf}}$ -based optimizer in the inner loop is articulated as
\begin{align}
\label{r_up}
    &\boldsymbol{(w,f)}^r=\boldsymbol{(w,f)}^{r-1}+\text{LSTM}_{\boldsymbol{wf}}\left(\nabla \hat{g}(\boldsymbol{(wf)}^{r-1}),\boldsymbol{C}_{\boldsymbol{(wf)}^{r-1}}, \boldsymbol{\theta}_{\boldsymbol{(wf)}}\right),
\end{align}
where $r=1,...,R$ is the $r$-th iteration for updating $\hat{g}(\boldsymbol{w})$ and $\hat{g}(\boldsymbol{f})$ in the inner loop. To strictly satisfy the constraints C4 and C5 in problem $\mathcal{P}$6, we use the projection method to transform the constraints into
\begin{equation}
{\boldsymbol{w}}=\begin{cases}{\boldsymbol{w}},\quad\text{if} \quad {\sum_{i=1}^{I}W_{i}}\leq W_{\text{RSU}}\\ 
\frac{{\boldsymbol{W}}}{\sum_{i=1}^{I} W_{i}}\sqrt{W_{\text{RSU}}},\quad \text{otherwise}\\
\end{cases},
\end{equation}

\begin{equation}
{\boldsymbol{f}}=\begin{cases}{\boldsymbol{f}},\quad\text{if} \quad {\sum_{i=1}^{I} f_{i}}\leq f_{\text{RSU}}\\ 
\frac{{\boldsymbol{f}}}{\sum_{i=1}^{I}f_{i}}\sqrt{f_{\text{RSU}}},\quad \text{otherwise}\\
\end{cases}.
\end{equation}



In the inner loop, the parameters of these four networks are fixed, which are used to generate the update of variables. As the input of the variable update function, the variables are updated once in each inner loop iteration, and the inner loop update of a subproblem is not completed until the number of updates reaches the set number of inner loop iterations. In this way, the four subproblems are updated iteratively with each other.

\subsubsection{The Network Parameters Updating in the Outer Loop}


In the outer loops, we update the network parameters through backpropagation to minimize the accumulated global loss, given by
\begin{align}
    \label{gloss}\mathcal{L}_{G}^{s}=\frac{1}{k_{\text{up}}}\sum_{k_s=(s-1)k_{\text{up}}+1}^{sk_{\text{up}}}G(\boldsymbol{\alpha}^{k_s}, \boldsymbol{\eta}^{k_s}, {\boldsymbol{p}}^{k_s},{\boldsymbol{w}}^{k_s},{\boldsymbol{f}}^{k_s}).
\end{align}
where $k_{up}$ is the update interval, and $s = 1, 2,..., S$, with $S=K/k_{up}$ being the maximum update number for LSTM networks and $K$ being the maximum outer steps. For every $k_{up}$ outer loop iteration, the parameters of the LSTM networks are updated by the Adam optimizer using the accumulated global loss $\mathcal{L}_{G}^{s}$. And the accumulated global loss $\mathcal{L}_{G}^{s}$ is used to update $\theta_{\boldsymbol{\alpha}}$, $\theta_{\boldsymbol{\eta}}$, $\theta_{\boldsymbol{p}}$ and $\theta_{\boldsymbol{wf}}$. Mathematically,
\begin{align}
     &\theta_{\boldsymbol{\alpha}}^{s+1} = \theta_{\boldsymbol{\alpha}}^{s} + \beta_{\boldsymbol{\alpha}} \cdot \text{Adam}(\theta_{\boldsymbol{\alpha}}^{s}, \nabla_{\theta_{\boldsymbol{\alpha}}^{s}}\mathcal{L}_{s}^{G}),\\
    &\theta_{\boldsymbol{\eta}}^{s+1} = \theta_{\boldsymbol{\eta}}^{s} + \beta_{\boldsymbol{\eta}} \cdot \text{Adam}(\theta_{\boldsymbol{\eta}}^{s}, \nabla_{\theta_{\boldsymbol{\eta}}^{s}}\mathcal{L}_{s}^{G}),\\
    &\theta_{\boldsymbol{p}}^{s+1} = \theta_{\boldsymbol{p}}^{s} + \beta_{\boldsymbol{p}}\cdot\text{Adam}(\theta_{\boldsymbol{p}}^{s}, \nabla_{\theta_{\boldsymbol{p}}^{s}}\mathcal{L}_{s}^{G}),\\
    &\theta_{\boldsymbol{wf}}^{s+1} =\theta_{\boldsymbol{wf}}^{s} + \beta_{\boldsymbol{wf}}\cdot\text{Adam}(\theta_{\boldsymbol{wf}}^{s}, \nabla_{\theta_{\boldsymbol{wf}}^{s}}\mathcal{L}_{s}^{G}),
\end{align}
where $\beta_{\boldsymbol{\alpha}}$, $\beta_{\boldsymbol{\eta}}, \beta_{\boldsymbol{p}}$ and $\beta_{\boldsymbol{wf}}$ are the learning rate of $\text{LSTM}_{\boldsymbol{\alpha}}$,  $\text{LSTM}_{\boldsymbol{\eta}}$, $\text{LSTM}_{\boldsymbol{p}}$ and $\text{LSTM}_{\boldsymbol{wf}}$, i.e., the iteration step size. The parameters $\theta $ are iteratively updated using the Adam method \cite{kingma2017adam}.

\subsubsection{The Overall Algorithm of the Proposed SKDML Method }

Algorithm 2 summarizes the proposed SKDML algorithm. Specifically, the algorithm starts with inputs of randomly initialized transmission mode selection $\boldsymbol{\alpha}_0$, task offloading ratio $\boldsymbol{\eta}_0$, bandwidth allocation $\boldsymbol{w}_0$, computational resource allocation $\boldsymbol{f}_0$, and randomly initialized network parameters $\theta_0$.  In the inner loop iteration, first, update the mode selection problem using LSTM networks to obtain the current optimal transmission mode.  The second, third and fourth inner-loop iterations are solved similarly to the first one by updating the task offloading ratio, the power selection and the allocation of the bandwidth and computing resources. In the outer-loop iterations, the network parameters are updated every $k_{up}$ inner-loop iteration after the global loss function is calculated $k_{up}$ times. The average loss function is used to update the network parameters of the four LSTM networks to customize the frequency of network parameter updates. By updating the networks in the inner loop and the outer loop, the optimal transmission mode selection, task offloading ratio, and resource allocation scheme are obtained.


\section{Numerical Results}
In this section, simulations are carried out to verify the effectiveness of the proposed traffic-aware task offloading mechanism.  The model-based AM algorithm and the data-driven meta-learning approach without knowledge are also conducted for comparison. Furthermore, the resource allocation with the conventional I-CC scheme is also considered as a baseline.

\begin{algorithm}[h]
\caption{The Proposed SKDML Method}
\begin{algorithmic}[1]
\STATE  Input: global loss function $\hat{G}({\boldsymbol{\alpha}^k,\boldsymbol{\eta}^k,\boldsymbol{p}}^{k},{\boldsymbol{w}}^{k},{\boldsymbol{f}}^{k})$, local loss functions $\hat{g}_({\boldsymbol{\alpha}})$, $\hat{g}_({\boldsymbol{\eta}})$, $\hat{g}_({\boldsymbol{p}})$ and $\hat{g}_({\boldsymbol{w,f}})$ , random initialization ${\boldsymbol{\eta}}^0$, ${\boldsymbol{p}}^0$, ${\boldsymbol{w}}^0$, ${\boldsymbol{f}}^0$, number of outer loops $K$, and number of inner loops $N$, $J$, $M$, $R$;
        \STATE Output: Estimated variables $\boldsymbol{\alpha}^K, \boldsymbol{\eta}^K, {\boldsymbol{p}}^K,{\boldsymbol{w}}^K,{\boldsymbol{f}}^K$;
        \FOR{$k\gets 0$ to $K$}
            \FOR{$n\gets 0$ to $N$}
                \STATE $\boldsymbol{\alpha}^n=\boldsymbol{\alpha}^{n-1}+\text{LSTM}_{\boldsymbol{\alpha}}(\nabla \hat{g}(\boldsymbol{\alpha}^{n-1}),\boldsymbol{C}_{\boldsymbol{\alpha}^{n-1}}, \boldsymbol{\theta}_{\boldsymbol{\alpha}})$;
            \ENDFOR    
            \STATE ${\boldsymbol{\alpha}}^k\gets {\boldsymbol{\alpha}}^N$;
            \STATE Update local loss function $\hat{g}_{{\boldsymbol{\eta}}^{k-1}, {\boldsymbol{p}}^{k-1}, {\boldsymbol{w}}^{k-1},{\boldsymbol{f}}^{k-1}}({\boldsymbol{\alpha^k}})$;
            \FOR{$j\gets 0$ to $J$}
                \STATE $\boldsymbol{\eta}^j=\boldsymbol{\eta}^{j-1}+\text{LSTM}_{\boldsymbol{\eta}}(\nabla \hat{g}(\boldsymbol{\eta}^{j-1}),\boldsymbol{C}_{\boldsymbol{\eta}^{j-1}}, \boldsymbol{\theta}_{\boldsymbol{\eta}}) $;
            \ENDFOR
            \STATE ${\boldsymbol{\eta}}^k\gets {\boldsymbol{\eta}}^J$
            \STATE Update local loss function $\hat{g}_{{\boldsymbol{\alpha}}^{k}, {\boldsymbol{p}}^{k-1}, {\boldsymbol{w}}^{k-1},{\boldsymbol{f}}^{k-1}}({\boldsymbol{\eta}^k})$;
            \FOR{$m\gets 0$ to $M$}
                \STATE $\boldsymbol{p}^m=\boldsymbol{p}^{m-1}+\text{LSTM}_{\boldsymbol{p}}(\nabla \hat{g}(\boldsymbol{p}^{m-1}),\boldsymbol{C}_{\boldsymbol{p}^{m-1}}, \boldsymbol{\theta}_{\boldsymbol{p}}) $;
            \ENDFOR
            \STATE ${\boldsymbol{p}}^k\gets {\boldsymbol{p}}^M$;
            \STATE Update local loss function $\hat{g}_{{\boldsymbol{\alpha}}^{k}, {\boldsymbol{\eta}}^{k}, {\boldsymbol{w}}^{k-1},{\boldsymbol{f}}^{k-1}}({\boldsymbol{p}^k})$;
            \FOR{$r\gets 0$ to $R$}
                \STATE $\boldsymbol{(w,f)}^r=\boldsymbol{(w,f)}^{r-1}+\text{LSTM}_{\boldsymbol{(w,f)}}(\nabla \hat{g}(\boldsymbol{(wf)}^{r-1}),\boldsymbol{C}_{\boldsymbol{(wf)}^{r-1}}, \boldsymbol{\theta}_{\boldsymbol{(wf)}})$;
            \ENDFOR
            \STATE ${\boldsymbol{w}}^k\gets {\boldsymbol{w}}^R$;
            \STATE ${\boldsymbol{f}}^k\gets {\boldsymbol{f}}^R$;
            \STATE Update local loss function $\hat{g}_{{\boldsymbol{\alpha}}^{k}, {\boldsymbol{\eta}}^{k}, {\boldsymbol{p}}^{k}}({\boldsymbol{w}}^{k}, \boldsymbol{f}^{k})$;
            \STATE Update global loss function $\hat{G}({\boldsymbol{\alpha}^k,\boldsymbol{\eta}^k,\boldsymbol{p}}^{k},{\boldsymbol{w}}^{k},{\boldsymbol{f}}^{k})$;
            \FOR{$s\gets 0$ to $K/{k_{\text{up}}}$}
                \STATE $\mathcal{L}_{G}^{s}=\frac{1}{k_{\text{up}}}\sum_{k_s=(s-1)k_{\text{up}}+1}^{sk_{\text{up}}}G(\boldsymbol{\alpha}^{k_s}, \boldsymbol{\eta}^{k_s}, {\boldsymbol{p}}^{k_s},{\boldsymbol{w}}^{k_s},{\boldsymbol{f}}^{k_s})$;
                \STATE $\theta_{\boldsymbol{\alpha}}^{s+1}=\theta_{\boldsymbol{\alpha}}^{s}-\beta_{\boldsymbol{\alpha}}\nabla_{\theta_{\boldsymbol{\alpha}}^s}\mathcal{L}_{G}^{s} $;
                \STATE $\theta_{\boldsymbol{\eta}}^{s+1}=\theta_{\boldsymbol{\eta}}^{s}-\beta_{\boldsymbol{\eta}}\nabla_{\theta_{\boldsymbol{\eta}}^s}\mathcal{L}_{G}^{s}$;
                \STATE $\theta_{\boldsymbol{p}}^{s+1}=\theta_{\boldsymbol{p}}^{s}-\beta_{\boldsymbol{p}}\nabla_{\theta_{\boldsymbol{P}}^s}\mathcal{L}_{G}^{s}$;
               \STATE $\theta_{{\boldsymbol{w}}{\boldsymbol{f}}}^{s+1}=\theta_{{\boldsymbol{w}}{\boldsymbol{f}}}^{s}-\beta_{{\boldsymbol{w}}{\boldsymbol{f}}}\nabla_{\theta_{{\boldsymbol{w}}{\boldsymbol{f}}}^s}\mathcal{L}_{G}^{s}$;
            \ENDFOR
        \ENDFOR
    \end{algorithmic}
\end{algorithm}

\begin{table}[htbp]
\caption{Simulation Parameters}
\begin{center}
\begin{tabular}{|c|c|c|c|}
\hline
\cline{1-2} 
 \textbf{\text{Parameter}}& \textbf{\text{Value}} \\
\hline
Bandwidth of RSUs & 40MHz \\
\hline
Transmit power vehicle & 300mW \\
\hline
Computing ability of RSUs & ${10}^{12}$ cycles/s \\
\hline
Maximum tolerant transmission latency & 100ms \\
\hline
Distance between RSUs & 500m \\
\hline

\end{tabular}
\label{tab1}
\end{center}
\end{table}
\subsection{Simulation Setup}
To validate the effectiveness of I-CSC mode and SKDML, we consider equipping each vehicle with an environmental perception task that exhibits divisibility, and its subtasks can be processed locally or offloaded to base stations for processing. We provide a detailed overview of the parameters and corresponding values used in this analysis, as outlined in Table 1 \cite{8758209}, \cite{8627987}, \cite{8649627}, \cite{8886130}.

We simulate and analyze the proposed algorithm based on Python. In the simulation, we consider a scenario where a BS accommodates ten vehicles, each carrying an environmental perception task. The ten vehicles are randomly positioned at distances of 300m, 400m, 500m, and 600m from the BS, respectively. We set the bandwidth of BS to 40MHz, and the transmit power of the vehicle is 300mW \cite{9439524}.

\subsection{Simulation Results}
The trend of the loss function convergence in AM algorithms, meta-learning without knowledge algorithms, and the proposed SKDML method in I-CSC mode, is illustrated in Fig. 4. The utilized neural network is LSTM. The gradient descent method's learning rate is set to $10^{-3}$, the positive size of AM algorithms $\beta$ is set to $8\times10^{-3}$ \cite{58337}. 


We set the outer loop iteration count of the proposed SKDML at 500 and the inner loop iteration count at 5. The convergence plots depict the proposed SKDML with a red line, the meta-learning without knowledge algorithms enhancement with a yellow line, and the AM algorithm with a blue line. From the graph, it is evident that the SKDML achieves convergence in nearly 100 iterations, the meta-learning without knowledge algorithm converges at around 150 iterations, whereas the AM algorithm exhibits gradual convergence, reaching its optimal state after 400 iterations. Its representations unequivocally demonstrate that the proposed SKDML algorithm outperforms the other two methods in terms of convergence speed. Furthermore, the meta-learning without knowledge algorithm showcases a superior convergence speed compared to the traditional AM algorithm. Adding to this, the loss value of the SKDML eventually converges to 80, whereas the loss of the meta-learning without knowledge algorithm converges to around 100, and the loss of the AM algorithm settles at approximately 200. This compellingly illustrates that the SKDML achieves a 60\% reduction in costs compared to the AM algorithm and a 50\% reduction compared to meta-learning without knowledge algorithm.
\begin{table}[htbp]
\caption{Online Processing Time of Three Algorithms}
\begin{center}
\begin{tabular}{|c|c|c|c|}
\hline
\cline{1-4} 
 \textbf{\text{Algorithm}}& \textbf{\text{Proposed SKDML method}} & \textbf{\text{Meta-learning w/o knowledge}} & \textbf{\text{AM algorithm}} \\
\hline
Time(ms) & 10.283943 & 19.586828 & 21.159225\\
\hline
\end{tabular}
\label{tab1}
\end{center}
\end{table}

In addition, we evaluated the convergence time of three algorithms, as presented in Table 2. Under identical environmental parameters, the SKDML demonstrated a convergence time of approximately 10 ms. This represents  47.2\% improvement in convergence speed compared to the meta-learning without knowledge algorithm  and  51.4\% improvement over AM algorithms. Furthermore, Table 2 presents the online inference time required by these three approaches, which is counted on a server with 11th Gen Intel(R) Core(TM) i7-11700 @ 2.50GHz, GPU: NVIDIA
GeForce RTX 3060.
\begin{figure}[htp]
    \centering
    \includegraphics[width=9cm]{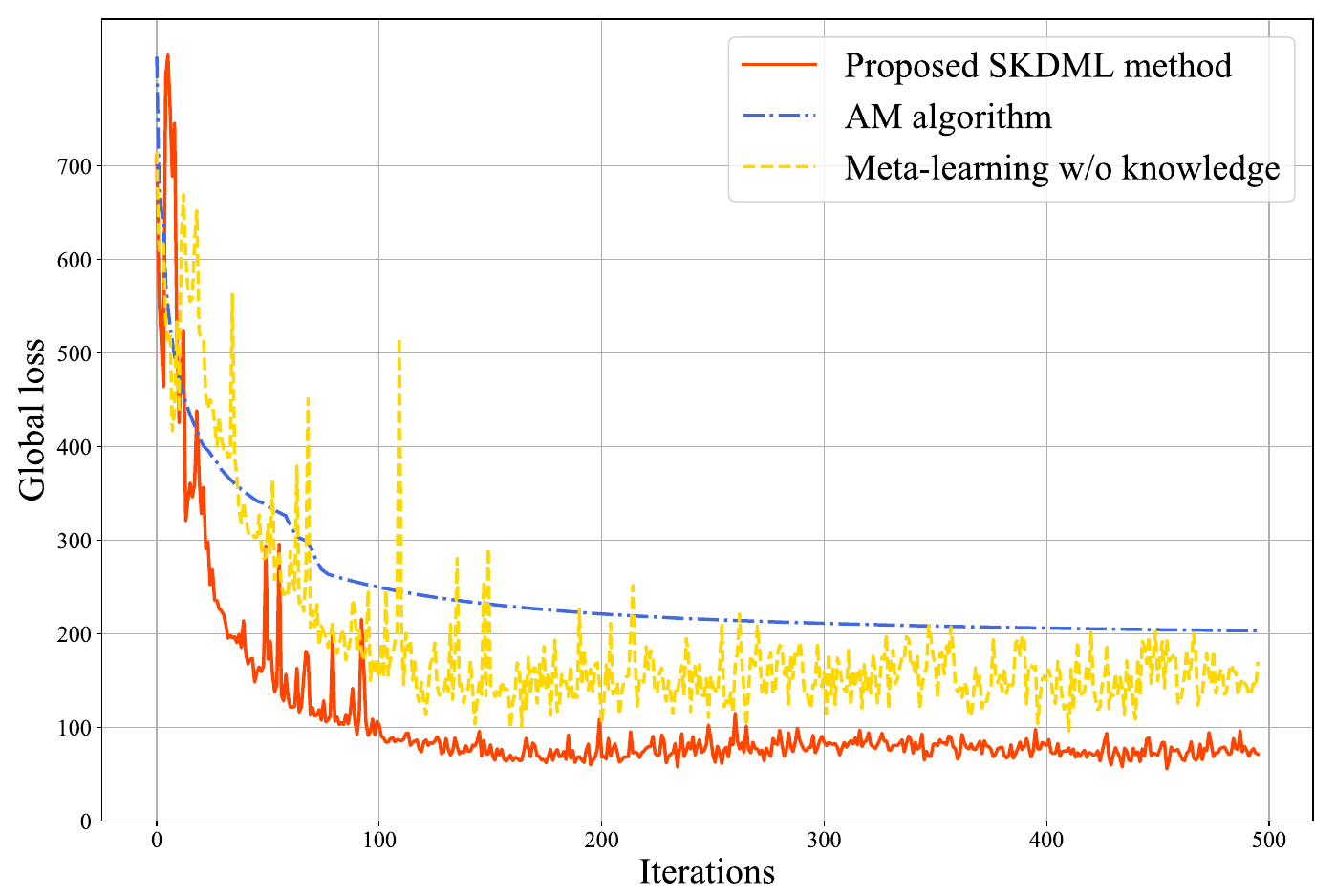}
    \caption{Convergence of the three algorithms}
    \label{fig:galaxy}
\end{figure}

\begin{figure}[htp]
    \centering
    \includegraphics[width=9cm]{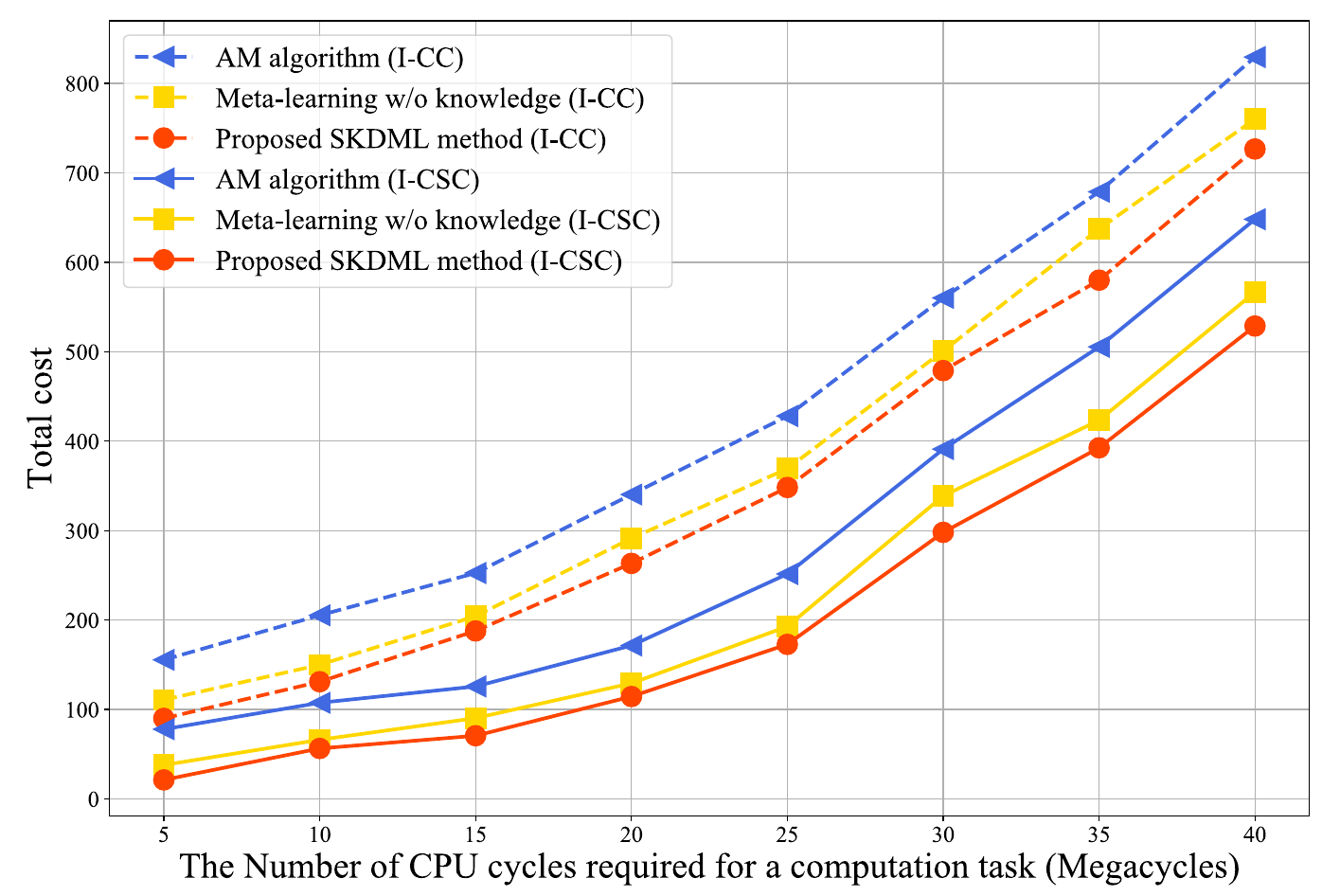}
    \caption{ Cost versus the resource required for task computing}
    \label{fig:galaxy}
\end{figure}
To compare the costs of three algorithms across various parameter variations, in the following, we studied the impact of different parameters (such as the task's required cycle count, latency tolerance, and the number of vehicles) on different algorithms and transmission methods.

In Fig. 5, we systematically vary the required cycle count for each task from 5 to 40 Megacycles \cite{8814220}. This variation enables a comprehensive assessment of the algorithms' performances as well as the associated costs linked to diverse transmission methods. As the required cycle count per task increases, both the costs of I-CC mode and I-CSC mode correspondingly escalate. Furthermore, the cost experiences a notably more pronounced escalation with the growing task cycle count. However, it's noteworthy that the cost of  I-CSC mode remains comparatively lower than that of I-CC mode. Among the three algorithms, the SKDML consistently demonstrates lower costs. This consistent trend suggests that, when the task cycle count is held constant, the SKDML facilitates environment-aware task completion with diminished costs, thereby yielding more efficient resource allocation strategies.

\begin{figure}[htp]
    \centering
    \includegraphics[width=9cm]{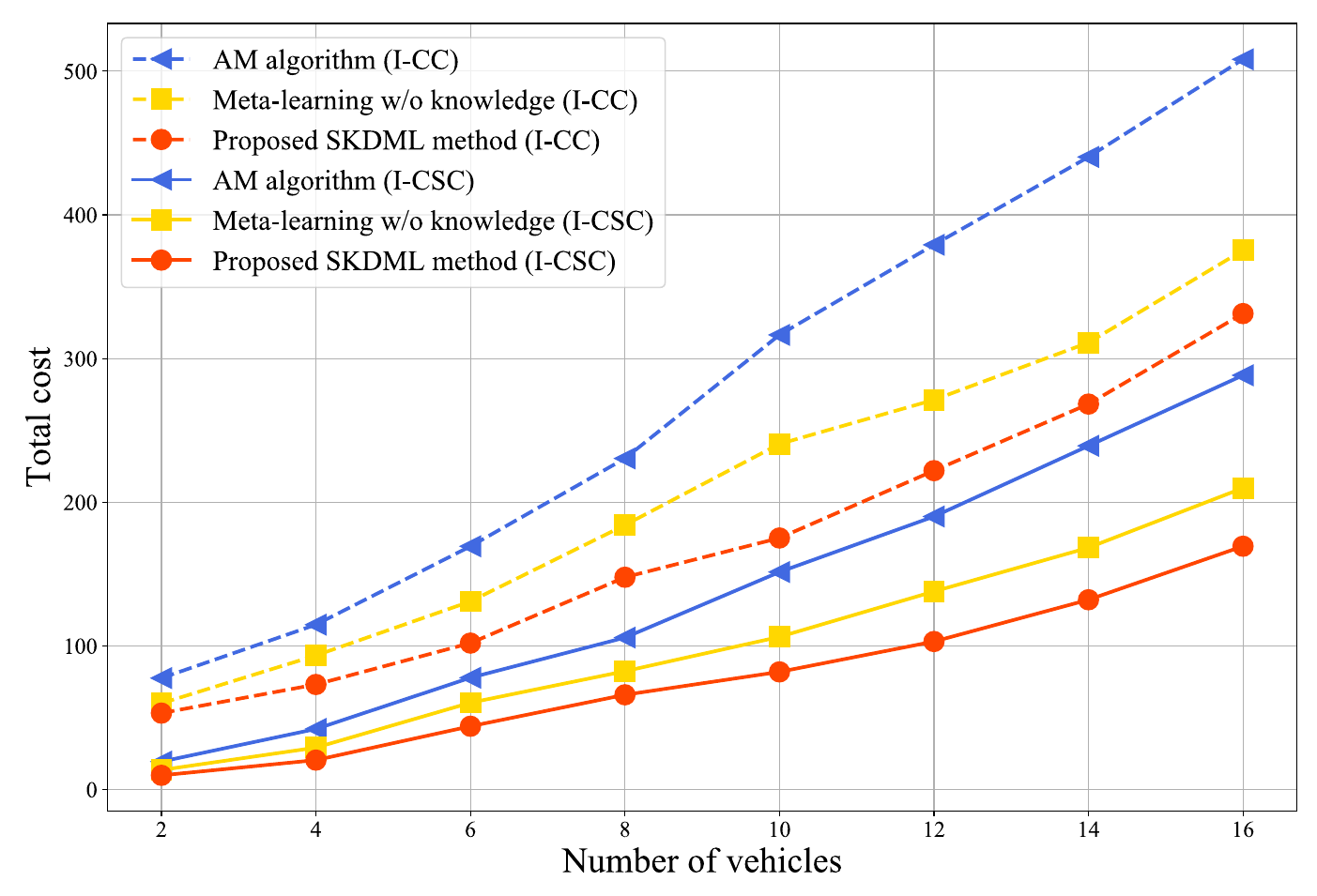}
    \caption{Cost versus the number of vehicies}
    \label{fig:galaxy}
\end{figure}

In Fig. 6, as the number of vehicles increases from 2 to 14, it becomes evident that the cost incurred by the I-CC mode consistently surpasses the cost associated with I-CSC mode. Additionally, the cost resulting from the AM algorithm exceeds that stemming from the meta-learning without knowledge algorithm, which in turn exceeds the cost originating from the SKDML. This pattern accentuates the notable efficacy of the SKDML. Given identical cost constraints, the SKDML demonstrates the capacity to accommodate a larger volume of vehicle tasks. In scenarios involving an equivalent number of vehicles, it engenders more streamlined resource allocation strategies.
\begin{figure}[htp]
    \centering
    \includegraphics[width=9cm]{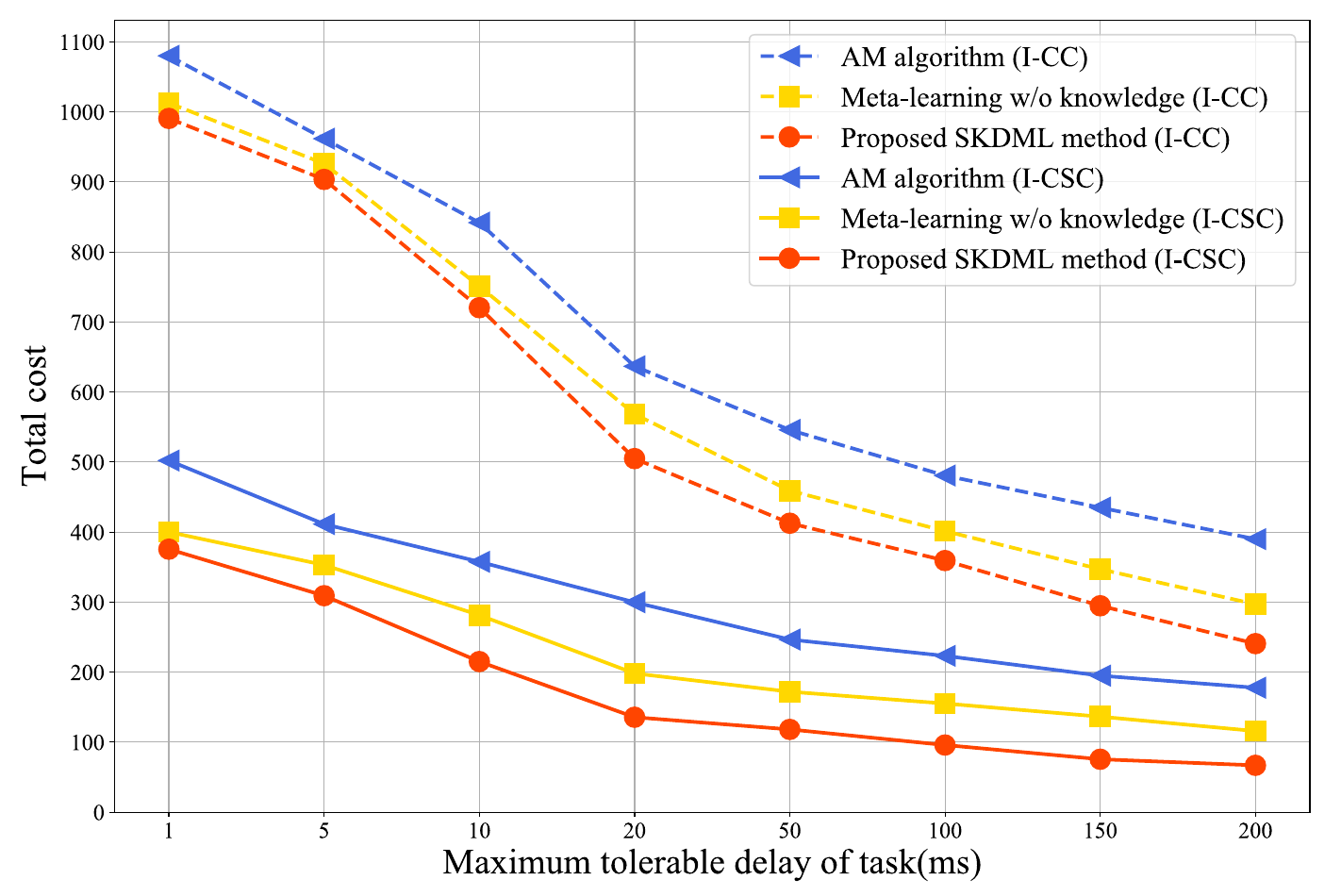}
    \caption{Cost versus the maximum tolerable latency of the task}
    \label{fig:galaxy}
\end{figure}

Latency tolerance for each task was systematically augmented from 1ms to 200ms, as depicted in Fig. 7. The figure elucidates that irrespective of variations in latency tolerance, the expense associated with I-CC mode consistently surpasses that linked to I-CSC mode. This observation underscores that when tasks share the same latency tolerance, opting for I-CSC mode can achieve task completion at a reduced cost. Furthermore, the SKDML consistently outperforms the other two algorithms.

To comprehensively validate disparities between the conventional I-CC mode and the I-CSC mode of transmission selection, this investigation expanded the data packet size from 1Mb to 30Mb, as illustrated in Fig. 8. When data packets are relatively small, ranging from 1Mb to 15Mb, the cost differential between the I-CC mode and the I-CSC mode, executed with the same algorithm, remains relatively inconspicuous. This outcome arises due to the I-CSC mode also opting for I-CC mode when dealing with diminutive data packets. 
However, as the data packet size escalates to 20Mb, the I-CSC mode transitions to the transmission of computational instructions. Due to the fact that packet size only affects a small portion of the total cost, the cost generated by I-CSC mode is much smaller than that of I-CC mode. Hence, post the 20Mb threshold, the cost of I-CSC mode progression slows down appreciably.

\begin{figure}[htp]
    \centering
    \includegraphics[width=9cm]{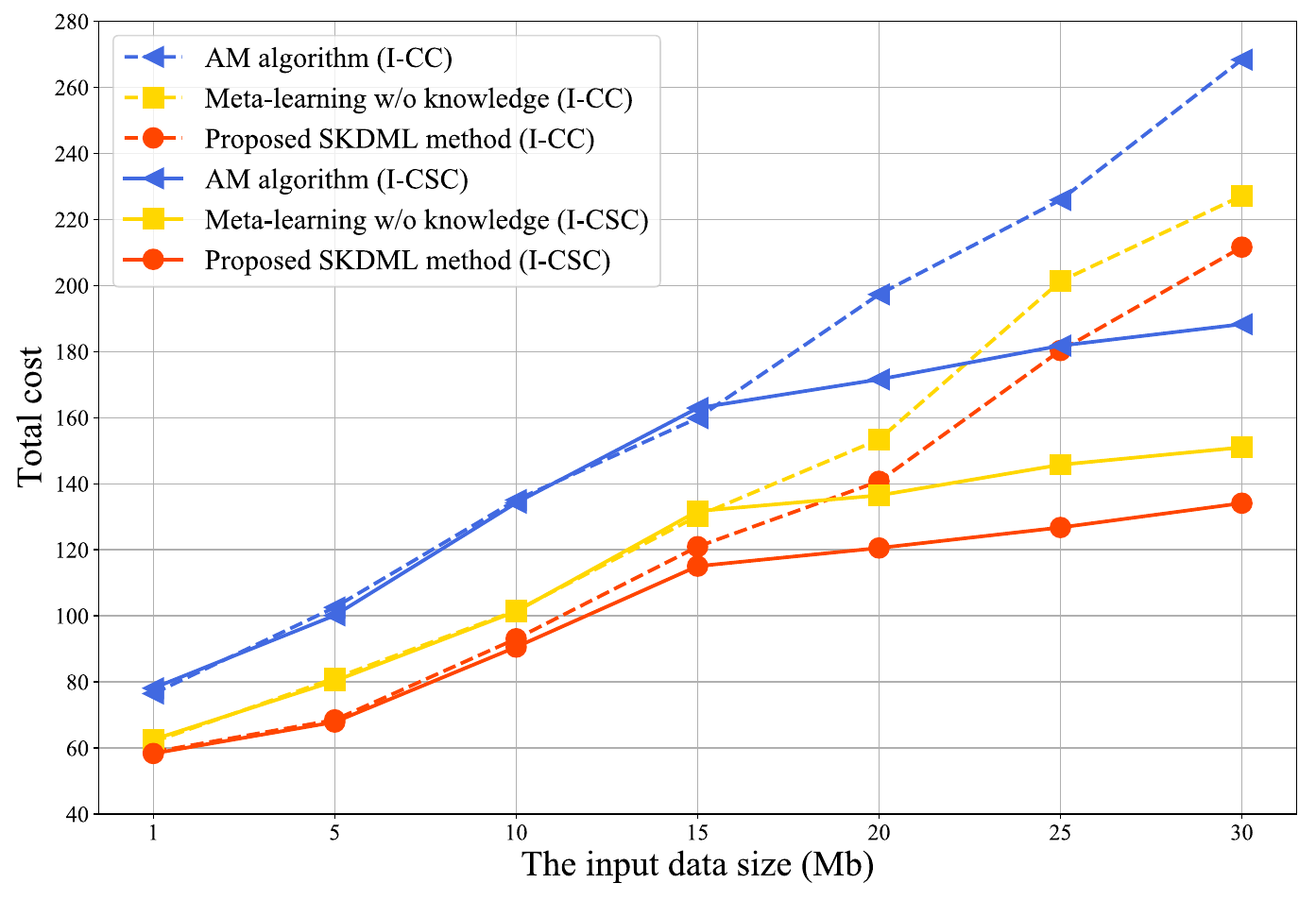}
    \caption{Cost versus the input data size of the task}
    \label{fig:galaxy}
\end{figure}

Fig. 9 presents bar graphs illustrating the performance of three algorithms under varying transmission methods as the data packet size ranges from 1Mb to 30Mb. The bar markers in distinct forms depict energy costs and payment costs in identical I-CC mode. These visualizations unveil that, as the data packet size transitions from 1Mb to 15Mb, energy costs and payment costs for the same algorithm under both transmission methods manifest substantial similarity. However, within the 15Mb to 30Mb range, the growth in energy costs and payment costs for I-CSC mode methods exhibits a more gradual trajectory compared to traditional I-CC mode. This pattern emerges due to the I-CC mode's inclination towards instruction-based transmission as data packet sizes surpass a certain threshold.

\begin{figure}[htp]

\includegraphics[width=8cm]{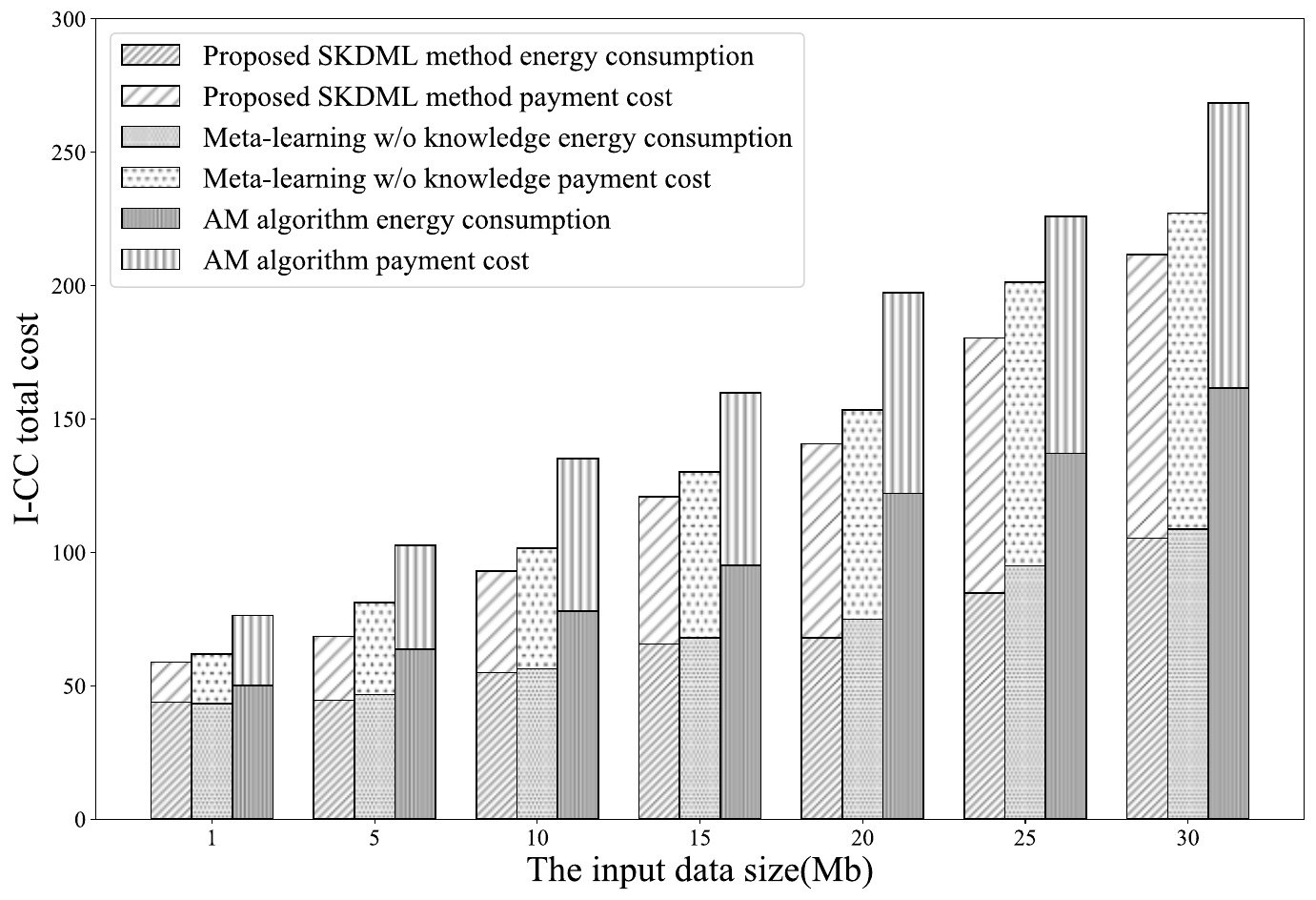}   
\includegraphics[width=8cm]{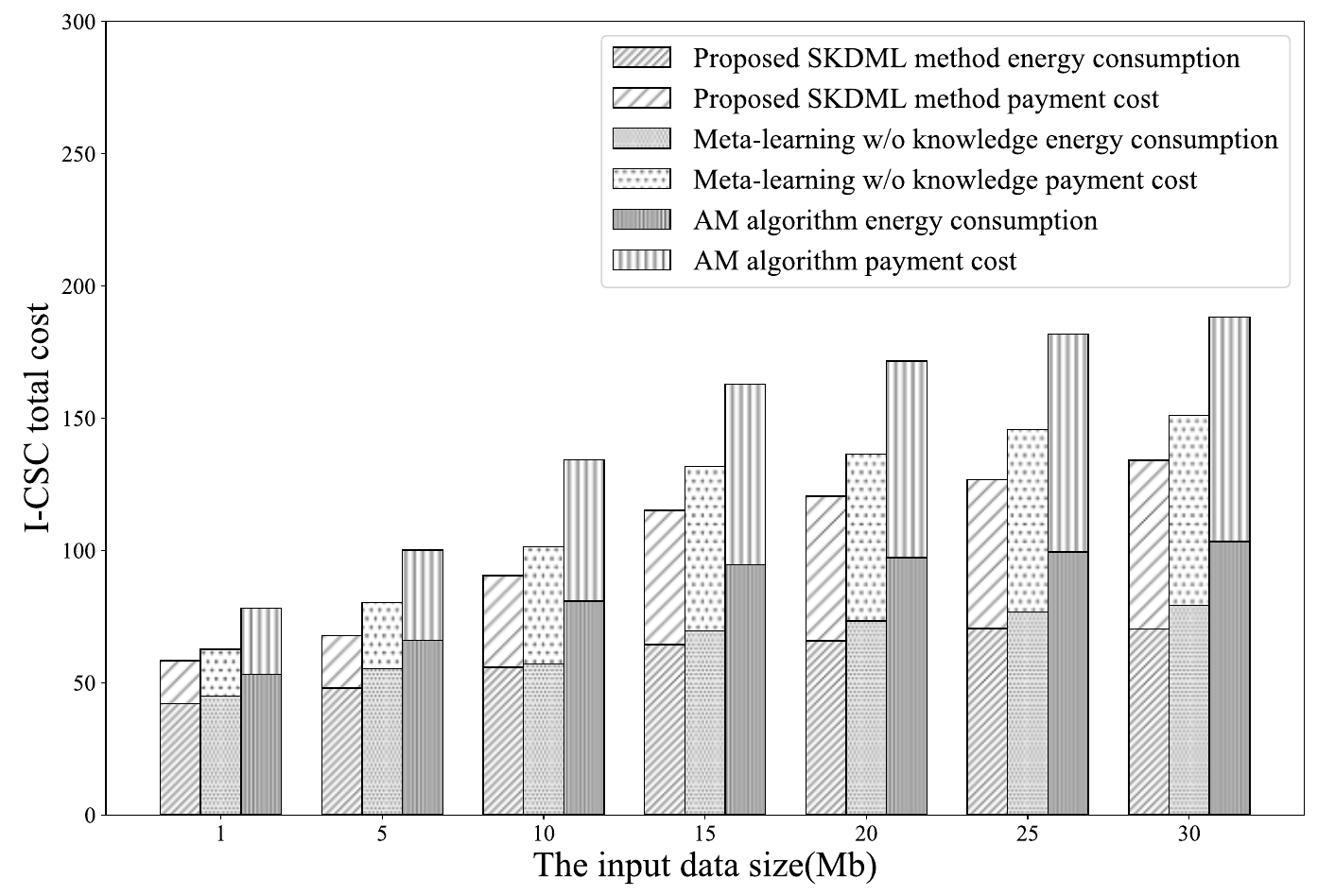}
\caption{Different transmission models versus the size of input data}
\label{1}
\end{figure}

\begin{figure}[htp]
    \centering
    \includegraphics[width=9cm]{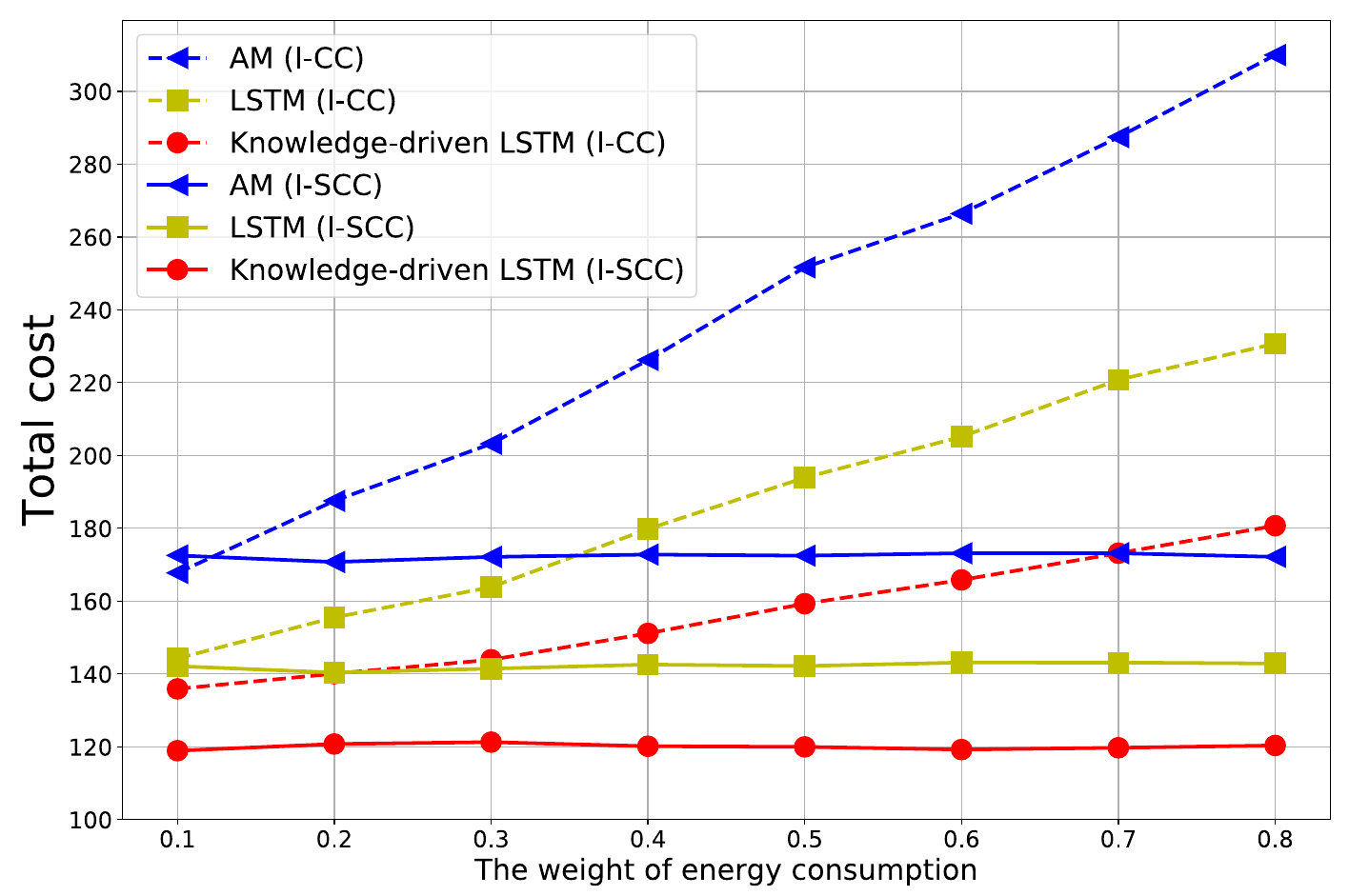}
    \caption{Total energy consumption] versus the weight of energy consumption in RSU.}
    \label{fig:galaxy}
\end{figure}

\section{Conclusion}

This paper has focused on the complex problem of computation offloading and resource allocation for environment-aware tasks. First, we have  introduced two distinct yet complementary offloading strategies: the conventional data packet offloading and the innovative environment-aware offloading. The primary objective of the optimization is to minimize the overall cost while adhering to the stringent constraints of task latency tolerance. Next, to augment real-time processing efficiency and interpretability, we have have proposed a novel approach called SKDML. It combines the well-established AM algorithm framework with insights distilled from neural networks. Simulation results have shown that our algorithm converges faster and performs better than AM algorithms and uninformed neural network methods. Moreover, even when different parameters are changed, the total cost obtained by our algorithm is consistently lower than that of the other two algorithms. The proposed algorithm not only surpasses the performance of both the traditional AM algorithm and meta-learning without knowledge algorithm in terms of convergence speed and overall efficacy but also exhibits exceptional performance in scenarios characterized by larger task data packet sizes. Not only that cost-effectiveness of environment-aware transmission methods has significantly outperformed that of conventional data packet transmission methods.


\bibliographystyle{elsarticle-num}
\bibliography{ref}

\begin{thebibliography}{10}
\expandafter\ifx\csname url\endcsname\relax
  \def\url#1{\texttt{#1}}\fi
\expandafter\ifx\csname urlprefix\endcsname\relax\def\urlprefix{URL }\fi
\expandafter\ifx\csname href\endcsname\relax
  \def\href#1#2{#2} \def\path#1{#1}\fi

\bibitem{8767077}
K.~Liu, X.~Xu, M.~Chen, B.~Liu, L.~Wu, V.~C.~S. Lee, A hierarchical
  architecture for the future internet of vehicles, IEEE Communications
  Magazine 57~(7) (2019) 41--47.

\bibitem{8744265}
S.~Liu, L.~Liu, J.~Tang, B.~Yu, Y.~Wang, W.~Shi, Edge computing for autonomous
  driving: Opportunities and challenges, Proceedings of the IEEE 107~(8) (2019)
  1697--1716.

\bibitem{8936542}
A.~Eskandarian, C.~Wu, C.~Sun, Research advances and challenges of autonomous
  and connected ground vehicles, IEEE Transactions on Intelligent
  Transportation Systems 22~(2) (2021) 683--711.

\bibitem{messous2017computation}
M.-A. Messous, H.~Sedjelmaci, N.~Houari, S.-M. Senouci, Computation offloading
  game for an {UAV} network in mobile edge computing, in: 2017 IEEE
  International Conference on Communications (ICC), IEEE, 2017, pp. 1--6.

\bibitem{8984345}
Y.~Wang, N.~Masoud, A.~Khojandi, Real-time sensor anomaly detection and
  recovery in connected automated vehicle sensors, IEEE Transactions on
  Intelligent Transportation Systems 22~(3) (2021) 1411--1421.

\bibitem{ren2019survey}
J.~Ren, D.~Zhang, S.~He, Y.~Zhang, T.~Li, A survey on end-edge-cloud
  orchestrated network computing paradigms: Transparent computing, mobile edge
  computing, fog computing, and cloudlet, ACM Computing Surveys (CSUR) 52~(6)
  (2019) 1--36.

\bibitem{li2020energy}
M.~Li, N.~Cheng, J.~Gao, Y.~Wang, L.~Zhao, X.~Shen, Energy-efficient
  uav-assisted mobile edge computing: Resource allocation and trajectory
  optimization, IEEE Trans. Veh. Technol. 69~(3) (2020) 3424--3438.

\bibitem{wang2022joint}
X.~Wang, L.~Fu, N.~Cheng, R.~Sun, T.~Luan, W.~Quan, K.~Aldubaikhy, Joint flying
  relay location and routing optimization for 6{G} {UAV--IoT} networks: A graph
  neural network-based approach, Remote Sensing 14~(17) (2022) 4377.

\bibitem{zhang2022federated}
Q.~Zhang, H.~Wen, Y.~Liu, S.~Chang, Z.~Han,
  Federated-reinforcement-learning-enabled joint communication, sensing, and
  computing resources allocation in connected automated vehicles networks, IEEE
  Internet Things J. 9~(22) (2022) 23224--23240.

\bibitem{8471165}
K.~Zhang, S.~Leng, X.~Peng, L.~Pan, S.~Maharjan, Y.~Zhang, Artificial
  intelligence inspired transmission scheduling in cognitive vehicular
  communications and networks, IEEE Internet of Things Journal 6~(2) (2019)
  1987--1997.

\bibitem{8555636}
J.~Zhou, F.~Wu, K.~Zhang, Y.~Mao, S.~Leng, Joint optimization of offloading and
  resource allocation in vehicular networks with mobile edge computing, in:
  2018 10th International Conference on Wireless Communications and Signal
  Processing (WCSP), 2018, pp. 1--6.

\bibitem{Dame_2013_CVPR}
A.~Dame, V.~A. Prisacariu, C.~Y. Ren, I.~Reid, Dense reconstruction using 3d
  object shape priors, in: Proceedings of the IEEE Conference on Computer
  Vision and Pattern Recognition (CVPR), 2013.

\bibitem{9439524}
Y.~Qi, Y.~Zhou, Y.-F. Liu, L.~Liu, Z.~Pan, Traffic-aware task offloading based
  on convergence of communication and sensing in vehicular edge computing, IEEE
  Internet of Things Journal 8~(24) (2021) 17762--17777.

\bibitem{5209891}
G.~Liu, W.~Wang, J.~Yuan, X.~Liu, Q.~Feng, Elimination of accumulated error of
  3d target location based on dual-view reconstruction, in: 2009 Second
  International Symposium on Electronic Commerce and Security, Vol.~2, 2009,
  pp. 121--124.

\bibitem{9982429}
Y.~Gong, Y.~Wei, Z.~Feng, F.~R. Yu, Y.~Zhang, Resource allocation for
  integrated sensing and communication in digital twin enabled internet of
  vehicles, IEEE Transactions on Vehicular Technology 72~(4) (2023) 4510--4524.

\bibitem{zhang2017optimal}
K.~Zhang, Y.~Mao, S.~Leng, S.~Maharjan, Y.~Zhang, Optimal delay constrained
  offloading for vehicular edge computing networks, in: 2017 IEEE International
  Conference on Communications (ICC), IEEE, 2017, pp. 1--6.

\bibitem{zhao2022adaptive}
H.~Zhao, J.~Tang, B.~Adebisi, T.~Ohtsuki, G.~Gui, H.~Zhu, An adaptive vehicle
  clustering algorithm based on power minimization in vehicular ad-hoc
  networks, IEEE Trans. Veh. Technol. 71~(3) (2022) 2939--2948.

\bibitem{8422240}
Y.~Liu, S.~Wang, J.~Huang, F.~Yang, A computation offloading algorithm based on
  game theory for vehicular edge networks, in: IEEE International Conference on
  Communications (ICC), 2018, pp. 1--6.

\bibitem{7726790}
H.~Shahzad, T.~H. Szymanski, A dynamic programming offloading algorithm for
  mobile cloud computing, in: IEEE Canadian Conference on Electrical and
  Computer Engineering (CCECE), 2016, pp. 1--5.

\bibitem{8543658}
J.~Du, F.~R. Yu, X.~Chu, J.~Feng, G.~Lu, Computation offloading and resource
  allocation in vehicular networks based on dual-side cost minimization, IEEE
  Transactions on Vehicular Technology 68~(2) (2019) 1079--1092.

\bibitem{8166725}
M.~Liu, Y.~Liu, Price-based distributed offloading for mobile-edge computing
  with computation capacity constraints, IEEE Wireless Communications Letters
  7~(3) (2018) 420--423.

\bibitem{7914660}
T.~Q. Dinh, J.~Tang, Q.~D. La, T.~Q.~S. Quek, Offloading in mobile edge
  computing: Task allocation and computational frequency scaling, IEEE
  Transactions on Communications 65~(8) (2017) 3571--3584.

\bibitem{7929399}
C.~Wang, C.~Liang, F.~R. Yu, Q.~Chen, L.~Tang, Computation offloading and
  resource allocation in wireless cellular networks with mobile edge computing,
  IEEE Transactions on Wireless Communications 16~(8) (2017) 4924--4938.

\bibitem{9112192}
M.~Cui, S.~Zhong, B.~Li, X.~Chen, K.~Huang, Offloading autonomous driving
  services via edge computing, IEEE Internet of Things Journal 7~(10) (2020)
  10535--10547.

\bibitem{dai2019artificial}
Y.~Dai, D.~Xu, S.~Maharjan, G.~Qiao, Y.~Zhang, Artificial intelligence
  empowered edge computing and caching for internet of vehicles, IEEE Wireless
  Commun. Mag. 26~(3) (2019) 12--18.

\bibitem{he2017integrated}
Y.~He, N.~Zhao, H.~Yin, Integrated networking, caching, and computing for
  connected vehicles: A deep reinforcement learning approach, IEEE Trans. Veh.
  Technol. 67~(1) (2017) 44--55.

\bibitem{9119487}
M.~Li, J.~Gao, L.~Zhao, X.~Shen, Deep reinforcement learning for collaborative
  edge computing in vehicular networks, IEEE Transactions on Cognitive
  Communications and Networking 6~(4) (2020) 1122--1135.

\bibitem{8297294}
M.~Cheng, J.~Li, S.~Nazarian, Drl-cloud: Deep reinforcement learning-based
  resource provisioning and task scheduling for cloud service providers, in:
  Asia and South Pacific Design Automation Conference (ASP-DAC), 2018, pp.
  129--134.

\bibitem{8676306}
Y.~Wang, H.~Liu, W.~Zheng, Y.~Xia, Y.~Li, P.~Chen, K.~Guo, H.~Xie,
  Multi-objective workflow scheduling with deep-q-network-based multi-agent
  reinforcement learning, IEEE Access 7 (2019) 39974--39982.

\bibitem{boyd2004convex}
S.~P. Boyd, L.~Vandenberghe, Convex optimization, Cambridge university press,
  2004.

\bibitem{9246287}
Q.~Hu, Y.~Cai, Q.~Shi, K.~Xu, G.~Yu, Z.~Ding, Iterative algorithm induced
  deep-unfolding neural networks: Precoding design for multiuser mimo systems,
  IEEE Transactions on Wireless Communications 20~(2) (2021) 1394--1410.

\bibitem{xia2022metalearning}
J.-Y. Xia, S.~Li, J.-J. Huang, Z.~Yang, I.~M. Jaimoukha, D.~G{\"u}nd{\"u}z,
  Metalearning-based alternating minimization algorithm for nonconvex
  optimization, IEEE Trans. Neural Netw. Learn. Syst. (2022).

\bibitem{NEURIPS2020_3d2d8ccb}
z.~luo, Y.~Huang, S.~Li, L.~Wang, T.~Tan, Unfolding the alternating
  optimization for blind super resolution, in: H.~Larochelle, M.~Ranzato,
  R.~Hadsell, M.~Balcan, H.~Lin (Eds.), Advances in Neural Information
  Processing Systems, Vol.~33, Curran Associates, Inc., 2020, pp. 5632--5643.

\bibitem{kingma2017adam}
D.~P. Kingma, J.~Ba, Adam: A method for stochastic optimization (2017).
\newblock \href {http://arxiv.org/abs/1412.6980} {\path{arXiv:1412.6980}}.

\bibitem{8758209}
I.~Sorkhoh, D.~Ebrahimi, R.~Atallah, C.~Assi, Workload scheduling in vehicular
  networks with edge cloud capabilities, IEEE Transactions on Vehicular
  Technology 68~(9) (2019) 8472--8486.

\bibitem{8627987}
Y.~Sun, X.~Guo, J.~Song, S.~Zhou, Z.~Jiang, X.~Liu, Z.~Niu, Adaptive
  learning-based task offloading for vehicular edge computing systems, IEEE
  Transactions on Vehicular Technology 68~(4) (2019) 3061--3074.

\bibitem{8649627}
P.~Liu, J.~Li, Z.~Sun, Matching-based task offloading for vehicular edge
  computing, IEEE Access 7 (2019) 27628--27640.

\bibitem{8886130}
K.~Zhang, X.~Gui, D.~Ren, Joint optimization on computation offloading and
  resource allocation in mobile edge computing, in: IEEE Wireless
  Communications and Networking Conference (WCNC), 2019, pp. 1--6.

\bibitem{58337}
P.~Werbos, Backpropagation through time: what it does and how to do it,
  Proceedings of the IEEE 78~(10) (1990) 1550--1560.

\bibitem{8814220}
J.-S. Lee, T.-H. Park, Fast lidar - camera fusion for road detection by cnn and
  spherical coordinate transformation, in: 2019 IEEE Intelligent Vehicles
  Symposium (IV), 2019, pp. 1797--1802.

\end{thebibliography}


\end{document}